 \documentclass[final,5p,times,twocolumn]{elsarticle}

\usepackage{amsmath}
\usepackage{amssymb}
\usepackage{amsthm}

\usepackage{pdfpages}

\usepackage{color}
\usepackage[bookmarks=false]{hyperref}
\usepackage{comment}

\newcommand{\tp}{{\scriptstyle\mspace{-1mu}\top}}
\DeclareMathAlphabet{\mathitbf}{OT1}{ptm}{bx}{it} 
\newcommand{\mat}[1]{\mathitbf{#1}} 
\newcommand{\oned}{\mbox{1-\textsc{d}}}
\newcommand{\twod}{\mbox{2-\textsc{d}}}
\newcommand{\threed}{\mbox{3-\textsc{d}}}
\newcommand{\mand}{\quad\text{and}\quad}
\newcommand{\mwhere}{\quad\text{where}\quad}
\newcommand{\nofbox}{}
\newcommand{\tof}{\textsc{tof}}
\newcommand{\LAMBDA}{\boldsymbol{\lambda}} 
\newcommand{\MU}{\boldsymbol{\mu}} 

\journal{Computer Vision and Image Understanding}

\begin{document}

\begin{frontmatter}



\title{Automatic Detection of Calibration Grids in Time-of-Flight Images}


\author[inria,qmul]{Miles Hansard}
\author[inria]{Radu Horaud\corref{cor1}}
\ead{Radu.Horaud@inria.fr}
\cortext[cor1]{Corresponding author}
\author[inria,supersonic]{Michel Amat}
\author[inria]{Georgios Evangelidis}
\address[inria]{INRIA Grenoble Rh\^{o}ne-Alpes, 38330 Montbonnot Saint-Martin, France}
\address[qmul]{School of Electronic Engineering and Computer Science, Queen Mary, University of London, Mile End Road, United Kingdom}
\address[supersonic]{SuperSonic Imagine, 510 Rue Ren\'e Descartes, 13857 Aix-en-Provence, France}

\begin{abstract}
It is convenient to calibrate time-of-flight cameras by established methods, using images of a chequerboard pattern. The low resolution of the amplitude image, however, makes it difficult to detect the board reliably. Heuristic detection methods, based on connected image-components, perform very poorly on this data. An alternative, geometrically-principled method is introduced here, based on the Hough transform. The projection of a chequerboard is represented by two pencils of lines, which are identified as oriented clusters in the gradient-data of the image. 
A projective Hough transform is applied to each of the two clusters, in axis-aligned coordinates. The range of each transform is properly bounded, because the corresponding gradient vectors are approximately parallel. Each of the two transforms contains a series of collinear peaks; one for every line in the given pencil. This pattern is easily detected, by sweeping a dual line through the transform. The proposed Hough-based method is compared to the standard OpenCV detection routine, by application to several hundred time-of-flight images. It is shown that the new method detects significantly more calibration boards, over a greater variety of poses, without any overall loss of accuracy. This conclusion is based on an analysis of both geometric and photometric error.
\end{abstract}


\end{frontmatter}
\newpage

\section{Introduction}
Time-of-flight (\tof{}) cameras \cite{hansard-2013} produce a \textit{depth image}, each pixel of which encodes the distance to the corresponding point in the scene. These devices emit pulsed infrared illumination, and infer distances from the time taken for light to reflect back to the camera. The \tof{} sensor can therefore be modelled, geometrically, as a pinhole device. Furthermore, knowledge of the \tof{} camera-parameters can be used to map raw depth-readings (i.e.~distances along lines of sight) into Euclidean scene-coordinates. The calibration thereby enables these devices to be used as stand-alone \threed\ sensors, or to be combined with ordinary colour cameras, for complete \threed\ modelling
and rendering~\cite{mure-2008,schiller-2008,zhu-2008,hahne-2009,koch-2009,kolb-2010,hansard-2011}.

\tof{} cameras can, in principle, be calibrated with any existing camera calibration method. For example, if a known chequerboard pattern is detected in a sufficient variety of poses, then the internal and external camera parameters can be estimated by standard routines
\cite{zhang-2000,hartley-2000,douskos-2008}.
It is possible to find the chequerboard vertices, in ordinary images,
by first detecting image-corners~\cite{harris-1988},
and subsequently imposing global constraints on their arrangement 
\cite{kruger-2004,wang-2007,bradski-2008}. This approach, however, is not
reliable for low-resolution images 
(e.g.~in the range $100$--$500\mspace{2mu}\mathrm{px}^2$) because the 
local image-structure is disrupted by sampling artefacts, as shown in 
fig.~\ref{fig:lowres}. Furthermore, these artefacts become worse as the 
board is viewed in distant and slanted positions, which are essential for
high quality calibration~\cite{datta-2009,lindner-2010}. \emph{The central motivation of this work is to detect a greater number and variety of calibration board-poses, in \tof{} images, without increasing the geometric error of the vertices}. The geometric error can be conveniently defined with respect to the known geometry of the board, as will be shown in section~\ref{sec:results}.

\tof{} sensors provide low-resolution depth and amplitude images. This is because relatively large detector-elements are required in order to allow accumulation of electrons, which increases the signal-to-noise ratio and yields accurate depth-estimates \cite{buttgen-2008} but, in turn, limits the spatial resolution of the devices. This explains 
the poor performance of heuristic detection-methods, when applied to \tof{} camera calibration. 
For example, the amplitude signal from a typical \textsc{\tof{}} camera
\cite{mesa-2010,foix-2011} resembles an ordinary greyscale image, but is of very
low spatial resolution (e.g.~176$\times 144$ for the SR4000 camera, 
or $160\times 120$ for the PMD PhotonICs on-chip sensor)
, as well as being noisy. 
A $712\times 496$ CMOS color-depth  sensor is currently being developed, but the resolution of the \tof{} image delivered by this sensor is only $356\times 248$ pixels \cite{kim-2012}. A $340 \times 96$ pixels \tof{} camera has also been developed, for driving applications \cite{niclass-2013}.
Lindner at al. \cite{lindner-2010} used the calibration module included in the OpenCV library \cite{bradski-2008} to estimate the parameters of a $200\times 200$ PMD depth-camera and noticed a high dependency between the intrinsic and extrinsic parameters. To overcome these issues, a calibration method that combines a \tof{} camera with colour cameras was proposed \cite{schiller-2008,lindner-2010}. While this method yields very accurate parameters, it requires a multiple-sensor setup composed of both \tof{} and standard cameras.

\begin{figure}[ht!]
\centering
\includegraphics[width=0.9\linewidth]{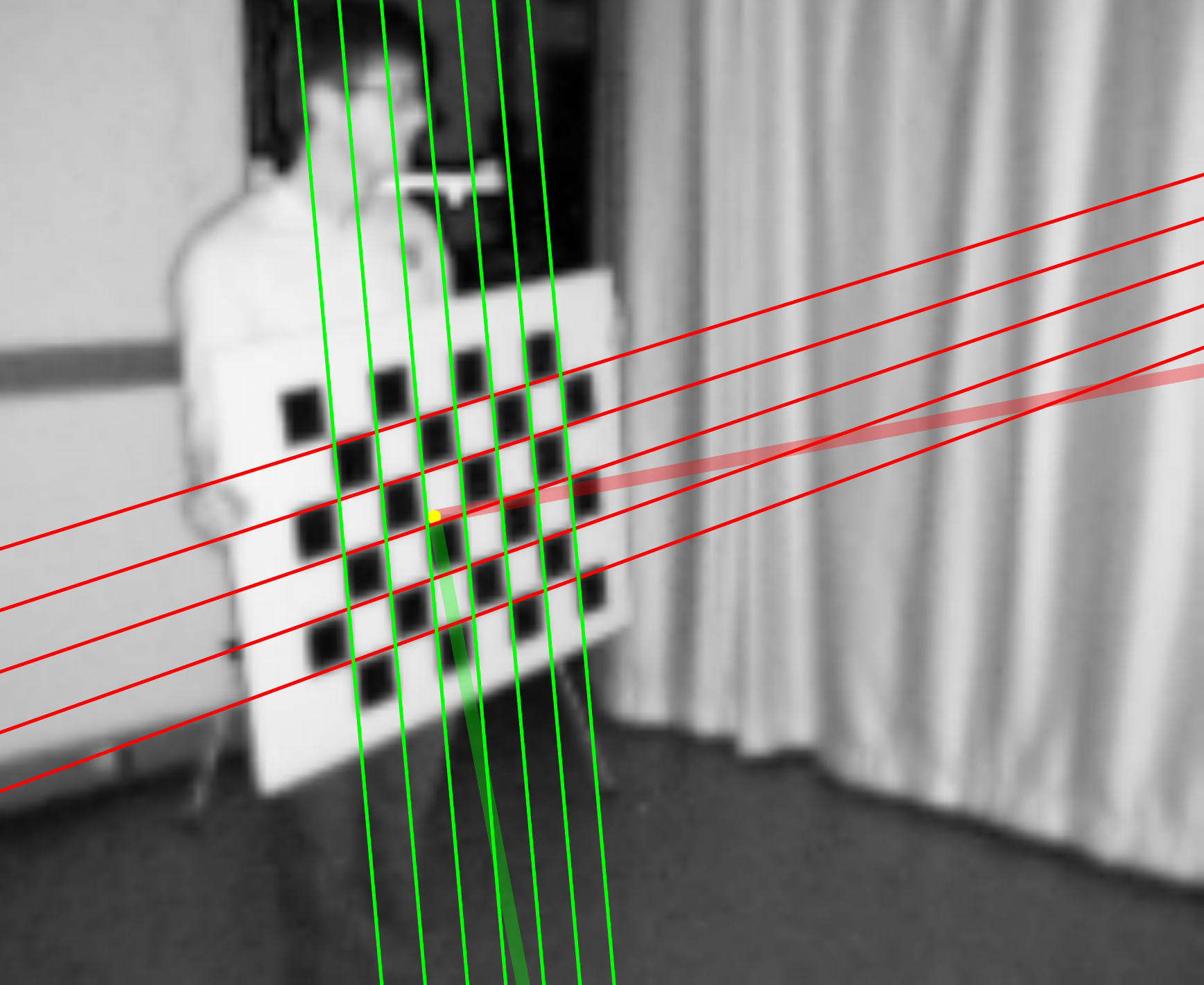}
\caption{An example of a \tof{} amplitude image taken in the infrared range in a normal neon-lit room. The original $176\times 144$ image has been magnified and smoothed for display.  The OpenCV software is unable to detect the vertices of the chequerboard in this case and in many other images as the board is viewed in distant and slanted positions. 
The red and green lines are the initial pencils detected by the proposed Hough-based method (prior to the refinement described in section \ref{sec:results}). The thick lines are the local Hough coordinate-systems,
which are automatically established, as described in section~\ref{sec:local-coordinates}.
}
\label{fig:example}
\end{figure}

Calibration grids are essentially composed of two pencils of lines, therefore chequerboard detection should explicitly take this structure into account. For instance, the method in \cite{ha-2007,ha-2009} starts by extracting points of interest, followed by eliminating those points that do not have a local chequerboard pattern, and finally by grouping together points lying along lines. This method puts a lot of emphasis on interest points, which are difficult to detect in low-resolution images, and does not take full advantage of the global structure of the calibration grid.

Two families of mutually orthogonal lines may also be detected by finding a dominant pair of vanishing-points. In \cite{barnard-1983} it is proposed to represent image lines on the Gaussian sphere (a unit sphere around the optical center of the camera). Under perspective projection, an image line projects onto a great circle on the Gaussian sphere, and a pencil of lines corresponds to a family great circles that intersect at antipodal points (see for example fig.~1 in \cite{shufelt-1999}). Therefore, a vanishing point may be found by detecting the intersections, provided that the camera's internal parameters are known. Vanishing point detection was implemented using a quantized Gaussian sphere and  a hierarchical (scale-space) Hough method, e.g.~\cite{quan-1989}. In general, Gaussian sphere-based methods require the detection of edges or of straight lines which are then projected as point sets (circles) on an azimuth-elevation grid, which may also produce spurious vanishing points~\cite{shufelt-1999}. 

More recently, vanishing-point detection was addressed as a clustering problem in the parameter space, using maximum likelihood and the EM algorithm \cite{kosecka-2006}, which requires suitable initialization. Alternatively, parameter-space clustering can be implemented using minimal sets \cite{toldo-2008} and random sampling. A method that combines \cite{toldo-2008} with an EM algorithm was recently proposed to find the three most orthogonal pencils of lines in indoor and outdoor scenes \cite{tardif-2009}. We tested this method using the software provided by the author\footnote{\url{http://www-etud.iro.umontreal.ca/~tardifj/fichiers/VPdetection-05-09-2010.tar.gz}} but found that the algorithm was not able to reliably extract and label edges from the low-resolution \tof{} amplitude images. 
We conclude that vanishing point methods, e.g., \cite{tardif-2009,antunes2013} fail to extract pencils of lines 
because they require accurate edge detection that is difficult to accomplish in low resolution, noisy images.



The method described in this paper is also based on the Hough transform~\cite{illingworth-1988}, 
but it effectively fits a specific model to the chequerboard pattern, e.g., fig.~\ref{fig:example}. This process is much less sensitive to
the resolution of the data, for two reasons. Firstly, information
is integrated across the source image, because each vertex
is obtained from the intersection of two fitted lines. Secondly, the
structure of a straight edge is inherently simpler than 
that of a corner feature.
However, for this approach to be viable, it is assumed that any lens distortion 
has been pre-calibrated, so that the images of
the pattern contain straight lines. This is not a serious 
restriction, because it is relatively easy to 
find enough boards (by any heuristic method) from which to obtain adequate estimates 
of the internal and lens parameters. Indeed there exist lens-calibration methods
that require only a single image~\cite{bukhari-2010,gonzalez-2011,melo2013}.
The harder problems of 
reconstruction and relative orientation can then be addressed 
after adding the newly detected boards, ending with a bundle-adjustment
that also refines the initial internal parameters.
Furthermore, the \textsc{\tof{}} devices used here have fixed lenses, which 
are sealed inside the camera body. This means that the internal and lens-distortion parameters
from previous calibrations can be re-used.

Another Hough-method for chequerboard detection has been presented by de la Escalera 
and Armingol~\cite{escalera-2010}. Their algorithm involves a \emph{polar}
Hough transform of all high-gradient points in the image. This results in
an array that contains a peak for each line in the pattern. It is not,
however, straightforward to extract these peaks, because their location
depends strongly on the unknown orientation of the image-lines. Hence 
all local maxima are detected by morphological operations, and a second
Hough transform is applied to the resulting data in~\cite{escalera-2010}. 
The true peaks will form two collinear sets in the first transform 
(cf.~sec.~\ref{sec:hough-transform}), and so the final task is to 
detect two peaks in the second Hough transform. This iteration makes it 
hard to determine an appropriate sampling scheme, and also increases
the computation and storage time of the procedure~\cite{tuytelaars-1997}.

The method described in this paper is quite different. It makes use 
of the gradient \emph{orientation} as well as magnitude at each point,
in order to establish an axis-aligned coordinate system for each image 
of the pattern. Separate Hough transforms are then performed in the
$x$ and $y$ directions of the local coordinate system. By construction,
the slope-coordinate of any line is close to zero in the corresponding 
\emph{Cartesian} Hough transform. This means that, on average, the peaks 
occur along a fixed axis of each transform, and can be detected by a simple
sweep-line procedure. Furthermore, the known $\ell\times m$ structure of 
the grid makes it easy to identify the optimal sweep-line in each transform. 
Finally, the two optimal sweep-lines map directly back to pencils of $\ell$ and 
$m$ lines in the original image, owing to the Cartesian nature of the 
transform. The principle of the method is shown in fig.~\ref{fig:overview}.

It should be noted that the method presented here was designed specifically
for use with \textsc{\tof{}} cameras. For this reason, the \emph{range}, as well as
intensity data are used to help segment the image in sec.~\ref{sec:preprocessing}.
However, this step could easily be replaced with an appropriate background 
subtraction procedure~\cite{bradski-2008}, in which case the new method could 
be applied to ordinary \textsc{rgb} images. Camera calibration is typically performed 
under controlled illumination conditions, and so there would be no need for
a dynamic background model.


\subsection{Overview and Contributions}

The new method is described in section~\ref{sec:method}; preprocessing
and segmentation are explained in sections~\ref{sec:preprocessing} and
\ref{sec:gradient-clustering} respectively, while section~\ref{sec:local-coordinates}
describes the geometric representation of the data. The necessary Hough transforms
are defined in section~\ref{sec:hough-transform}, and analyzed in sections~\ref{sec:hough-analysis}
and \ref{sec:decision-functions}.
The new method is evaluated on over~700 detections in section~\ref{sec:results}, 
and shown to be substantially better, for \textsc{\tof{}} images, than the standard OpenCV 
method. Conclusions are stated in section~\ref{sec:discussion}. \footnote{Supplementary material can be found at \url{http://www.eecs.qmul.ac.uk/~milesh/detect-suppl.pdf}}.

The main contributions of this paper are the use of a double-angle mapping
to segment the gradient vectors (\ref{sec:gradient-clustering}), 
the splitting of detection process into a pair of Cartesian Hough transforms
(\ref{sec:hough-transform}), and the sweep-line method of analyzing these
transforms (\ref{sec:hough-analysis}, \ref{sec:decision-functions}). 

\subsection{Notation}

Matrices and vectors will be written in bold, e.g.\ $\mat{M}$, $\mat{v}$, and
the Euclidean length of $\mat{v}$ will be written $|\mat{v}|$.
Equality up to an overall nonzero-scaling will be written
$\mat{v}\simeq\mat{u}$.
Image-points and lines will be represented in homogeneous coordinates~\cite{hartley-2000},
$\mat{p}\simeq (x,y,1)^\tp$ and $\LAMBDA \simeq (\alpha,\beta,\gamma)$,
such that $\LAMBDA\mspace{3mu}\mat{p} = 0$ if $\LAMBDA$ passes 
through $\mat{p}$. The intersection-point of two homogeneous
lines can be obtained
from the cross-product $(\LAMBDA\times\MU)^\tp$.
An assignment from variable $a$ to variable $b$ will be written $b\leftarrow a$. 
It will be convenient, for consistency with the pseudo-code listings, to use 
the notation $(m:n)$ for the sequence of integers from $m$ to $n$ inclusive. 
The `null' symbol $\varnothing$ will be used to denote undefined or unused variables
in the algorithms.

\section{Method}
\label{sec:method}

It is convenient to begin with an overview of the complete algorithm, before describing the exact form of the input data. Following this, subsections \ref{sec:preprocessing}--\ref{sec:decision-functions} will describe each step in detail. The individual stages of the algorithm are as follows:

\begin{enumerate}[\itshape A.]
\setlength{\itemsep}{3pt}
\item\emph{Preprocessing.}
The background of the image is roughly identified,
by depth-thresholding or image-differencing, and discarded. The gradient of
the remaining image is then computed.
\item\emph{Gradient Clustering.}
Two gradient clusters, one for each parallel set of edges
on the board, are identified. All weak gradients are discarded.
\item\emph{Local Coordinates.}
A pair of orthogonal axes, centred on the board, are constructed
from the two gradient clusters.
\item\emph{Hough Transform.}
Two Cartesian Hough transforms are performed, one for each gradient
cluster. The local coordinate system ensures that all edge
directions can be properly represented.
\item\emph{Hough Analysis.}
A line is swept across both transforms, until a collinear set 
of peaks is found in each case. The two sets of peaks correspond 
to two pencils of image-lines, the intersections of which are the 
estimated vertices of the calibration grid.
\item\emph{Decision Functions.}
The solution is accepted if two tests are passed: Firstly, the separation
of adjacent lines in each pencil must not be too variable. Secondly, there must 
be black squares on both sides of the outermost lines of each pencil.
\end{enumerate}

\begin{figure}[t!]
\centering
\begin{tabular}{cc}
\includegraphics[width=0.49\linewidth]{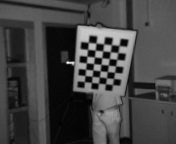} &
\includegraphics[width=0.49\linewidth]{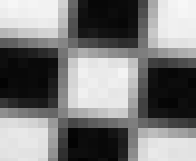}
\end{tabular}
\caption{Example chequers from a \textsc{\tof{}} amplitude image (shown on the left).
Note the variable appearance of the four junctions at this resolution; \mbox{`$\times$ like'} at lower-left vs.\ \mbox{`$+$ like'} at top-right.}
\label{fig:lowres}
\end{figure}

\begin{figure}[t!]
\centering
\includegraphics[width=0.8\linewidth]{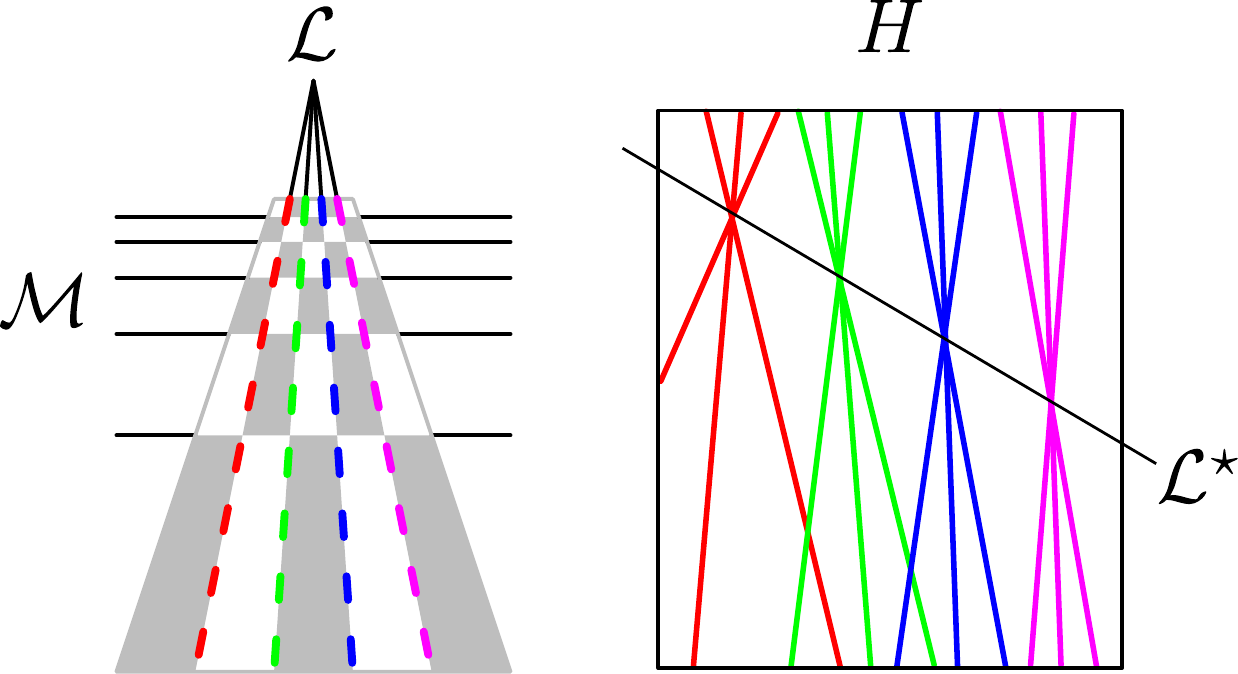}
\caption{\textbf{Left:} A perspective image of a calibration grid is represented
by line-pencils $\mathcal{L}$ and $\mathcal{M}$, which intersect at the $\ell\times m = 20$
internal vertices of this board. 
Strong image-gradients are detected along the dashed lines. 
\textbf{Right:} The Hough transform $H$ of the image-points associated with 
$\mathcal{L}$. Each high-gradient point maps to a line, such that there 
is a pencil in $H$ for each set of edge-points. The line $\mathcal{L}^\star$,
which passes through the $\ell=4$ Hough-vertices, is the Hough representation of
the image-pencil $\mathcal{L}$.}
\label{fig:overview}
\end{figure}

The form of the input data will now be detailed.
Suppose that the chequerboard has $(\ell+1) \times (m+1)$ squares,
with $\ell < m$. It follows that the \emph{internal} vertices of the 
pattern are imaged as the $\ell m$ line-intersections
\begin{equation}
\mat{v}_{ij} = \LAMBDA_i\times\MU_j
\quad\text{where}\
\begin{cases}
\mbox{$\LAMBDA_i \in \mathcal{L}$ for $i = 1:\ell$, and}\\
\mbox{$\MU_j \in \mathcal{M}$ for $j = 1:m$}.
\end{cases}
\label{eqn:vertices}
\end{equation}
The sets $\mathcal{L}$ and $\mathcal{M}$ are \emph{pencils}, meaning that the
$\LAMBDA_i$ all intersect at a point $\mat{p}$, while the
$\MU_j$ all intersect at a point $\mat{q}$. Note that 
$\mat{p}$ and $\mat{q}$ are the \emph{vanishing points} of the grid-lines,
which may be at infinity in the images (cf.~pencil~$\mathcal{M}$ in fig.~\ref{fig:overview}).

It is assumed that the imaging device, such as a \textsc{\tof{}} camera, provides
a range map $D_{xy}$, containing distances from the optical centre, as well
as a luminance-like amplitude map $A_{xy}$, where $(x,y)$ represents a pixel. The images $D$ and $A$ are of size $X\times Y$. All images must be undistorted, as described in the 
introduction.

\subsection{Preprocessing}
\label{sec:preprocessing}

The amplitude image $A$ is roughly segmented, by discarding all pixels that
correspond to very near or far points. This gives a new image $B$, which 
typically contains the board, plus the person holding it:
\begin{equation}
B_{xy} \leftarrow
\begin{cases}
\mbox{$A_{xy}$ \text{if} $d_0 < D_{xy} < d_1$}\\ 
\mbox{$\varnothing$ otherwise}.
\end{cases}
\label{eqn:amp-segm}
\end{equation}
The near-limit $d_0$ is determined by the closest position for which
the board remains fully inside the field-of-view of the camera. The 
far-limit $d_1$ is typically set to a value just closer than the far
wall of the scene. These parameters need only be set approximately, 
provided that the interval $[d_1, d_0]$ covers the possible positions of
the calibration board.

It is useful to perform a morphological erosion operation 
at this stage, in order to partially remove the perimeter of the board.
In particular, if the physical edge of the board is not white, then it 
will give rise to irrelevant image-gradients.
The erosion radius need only be set approximately (a value of 2px was used here), assuming
that there is a reasonable amount of white-space around the chessboard
pattern.

The gradient of the remaining amplitude image is now computed, using the
simple kernel $\Delta = (-\mbox{1/2},\, \mbox{0},\,\mbox{1/2})$.
The horizontal and vertical components are
\begin{equation}
\begin{aligned}
\xi_{xy} &\leftarrow (\Delta \star B)_{xy}\\
&=\rho\cos\theta 
\end{aligned}
\qquad\text{and}\qquad
\begin{aligned}
\eta_{xy} &\leftarrow (\Delta^{\!\tp} \!\star B)_{xy}\\
&=\rho\sin\theta
\end{aligned}
\label{eqn:grad}
\end{equation}
where $\star$ indicates convolution, and $(\rho,\theta)$ is the polar representation of the gradient.
No pre-smoothing of the image is performed, owing to the 
low spatial resolution of the data. 

\subsection{Gradient Clustering}
\label{sec:gradient-clustering}

The objective of this section is to assign each gradient vector $(\xi_{xy},\eta_{xy})$
to one of three classes, with labels $\kappa_{xy}\in\{\lambda$, $\mu$, $\varnothing$\}.
If $\kappa_{xy} = \lambda$ then pixel $(x,y)$ is on one of the lines in $\mathcal{L}$,
and $(\xi_{xy},\eta_{xy})$ is perpendicular to that line. If $\kappa_{xy} = \mu$, then
the analogous relations hold with respect to $\mathcal{M}$. If $\kappa_{xy}=\varnothing$
then pixel $(x,y)$ does not lie on any of the lines.

\begin{figure}[t!]
\centering
\includegraphics[width=.45\linewidth]{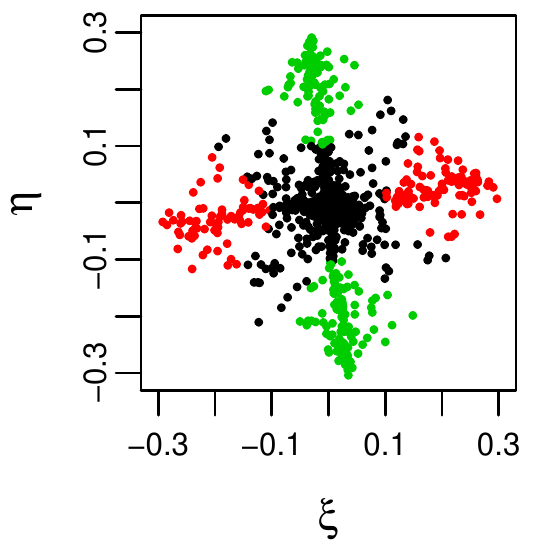}\hspace{.5em}
\includegraphics[width=.45\linewidth]{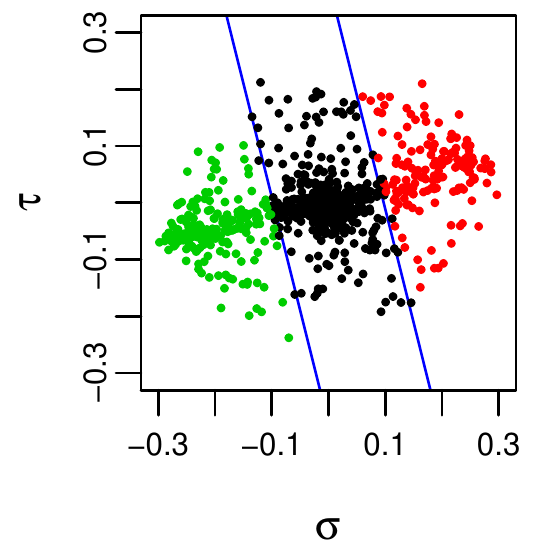}\hspace{.5em}
\caption{\textbf{Left:} the cruciform distribution of image gradients, due to black/white 
and white/black transitions at each orientation, would be difficult to segment in
terms of horizontal and vertical components $(\xi,\eta)$.
\textbf{Right:} the same distribution is easily segmented, by eigen-analysis, in the 
double-angle representation of equation~(\ref{eqn:double-angle}). The red and green labels 
are applied to the corresponding points in the
original distribution, on the left.}
\label{fig:scatter}
\end{figure}

The gradient distribution, after the initial segmentation, will contain two elongated 
clusters through the origin, which will be approximately orthogonal.
Each \emph{cluster} corresponds to a gradient orientation (mod~$\pi$), while each \emph{end} of
a cluster corresponds to a gradient polarity (black/white vs.\ white/black).
Two methods of identifying these clusters are described below; a principal component method, and a \textsc{ransac} method.


The \emph{principal component method} begins with a double-angle mapping of the data~\cite{granlund-1978},
which will be expressed as $(\xi,\eta) \mapsto (\sigma,\tau)$.
This mapping results in a \emph{single} elongated cluster, each \emph{end} of which corresponds
to a gradient orientation (mod~$\pi$). A real example of this mapping is shown in fig.~\ref{fig:scatter}.

The double-angle coordinates are obtained by applying the trigonometric identities $\cos(2\theta) = \cos^2\theta - \sin^2\theta$ and 
$\sin(2\theta) = 2\sin\theta\cos\theta$ to the gradients (\ref{eqn:grad}), so that
\begin{gather}
\sigma_{xy} \leftarrow \frac{1}{\rho_{xy}}\bigl(\xi_{xy}^2 - \eta_{xy}^2\bigr) 
\mand
\tau_{xy} \leftarrow \frac{2}{\rho_{xy}} \mspace{1mu}\xi_{xy} \mspace{2mu} \eta_{xy} \label{eqn:double-angle}\\[1ex]
\mwhere \rho_{xy} = \sqrt{\xi_{xy}^2 + \eta_{xy}^2} \nonumber
\end{gather}
for all points at which the magnitude $\rho_{xy}$ is above machine precision.
Let the first unit-eigenvector of the $(\sigma,\tau)$ covariance matrix be 
$\bigl(\cos(2\phi)\bigr.$, $\bigl.\sin(2\phi)\bigr)$, which is written in this way so that 
the angle $\phi$ can be interpreted in the original image. The cluster-membership is now defined by the 
orthogonal projection 
\begin{equation}
\pi_{xy} = \bigl(\sigma_{xy},\, \tau_{xy}\bigr) \cdot \bigl(\cos(2\phi),\, \sin(2\phi)\bigl)
\label{eqn:axis}
\end{equation}
of the data onto this axis. It is now straightforward to classify the gradient-vectors $(\xi_{xy},\eta_{xy})$, as shown in fig.~\ref{fig:scatter}, according to a threshold $\pi_{\min}$;
\begin{equation}
\kappa_{xy} \leftarrow 
\begin{cases}
\lambda &\text{if } \pi_{xy} \ge \pi_{\min} \\
\mu     &\text{if } \pi_{xy} \le -\pi_{\min} \\
\varnothing &\text{otherwise}.
\end{cases}
\label{eqn:labelling}
\end{equation}
Hence the symbol $\varnothing$ is assigned to all strong gradients that are \emph{not} aligned with either axis of the board, as well as to \emph{all} weak gradients. 

The above method is robust to moderate perspective effects, because only the first principal component is needed, and this is well-defined for any elongated distribution. However, in order to include boards with extreme perspective distortion, an even more robust \textsc{ransac} method can be used, as described below.

The \emph{{\footnotesize RANSAC}~method} is based on the fact that two lines through the origin can be defined by two points; one on each line. These two points are randomly sampled in the $(\xi,\eta)$ gradient space, and used to define two normal vectors 
$(-\eta_\lambda,\xi_\lambda)$ and 
$(-\eta_\mu,\xi_\mu)$.
The projections of the all $(x,y)$ gradient points, onto each unit vector, 
are defined as
\begin{align}
\pi_{xy}^\lambda &= \frac{(-\eta_\lambda, \xi_\lambda) \cdot(\xi_{xy},\eta_{xy})}{\sqrt{\xi_\lambda^2+\eta_\lambda^2}} \\[1ex]
\pi_{xy}^\mu &= \frac{(-\eta_\mu, \xi_\mu) \cdot(\xi_{xy},\eta_{xy})}{\sqrt{\xi_\mu^2+\eta_\mu^2}}
\end{align}
The gradients are are now classified, in relation to a slab of thickness $2\times\pi_{\min}$ around each line:
\begin{equation}
\kappa_{xy} \leftarrow
\begin{cases}
\lambda &\text{if}\,\,\, |\pi_{xy}^\lambda| \le \pi_{\min} \,\,\,\text{and}\,\,\, |\pi_{xy}^\mu| > \pi_{\min}\\
\mu &\text{if}\,\,\, |\pi_{xy}^\mu| \le \pi_{\min} \,\,\,\text{and}\,\,\, |\pi_{xy}^\lambda| > \pi_{\min}\\
\varnothing &\text{otherwise}.
\end{cases}
\end{equation}
Points that are in \emph{both} slabs (i.e.~around the intersection) are given the null label, which means that more points are excluded as the two lines become more parallel. This has desirable effect of automatically excluding more weak gradients as the perspective distortion increases.
The quality of the classification is defined as the number of non-null labels, and the best solution is chosen, as usual, from the ensemble of samples.

In general, the \textsc{ransac} method is more robust that the the principal component method; it does, however, have two drawbacks. Firstly, it is necessary to set the number of random samples to be drawn, which introduces an additional parameter. Secondly, the final result of the calibration will not be exactly repeatable, unless the same random-number generator and seed are employed each time.


Suppose that the gradients have now been classified, using either the principal component method or the \textsc{ransac} method, as described above. The next task is to resolve the respective \emph{identities} of the clusters, with respect to labels $\lambda$ and $\mu$. 
In principle, the class that contains the 
greater number of gradients should correspond to $\mathcal{L}$, which was defined as the pencil containing
fewer lines. This is because, for a fronto-parallel board, the total \emph{lengths} of 
the edges in $\mathcal{L}$ and $\mathcal{M}$ are proportional to $\ell(m-1)$ and $m(\ell-1)$ 
respectively, with $\ell<m$. This prediction is unreliable, in practice, owing to 
foreshortening and other image-effects. For this reason, the correspondence 
$\{\lambda,\mu\} \Leftrightarrow \{\mathcal{L}, \mathcal{M}\}$ between labels
and pencils will be resolved more robustly, in section~\ref{sec:hough-analysis}.

\subsection{Local Coordinates}
\label{sec:local-coordinates}

A coordinate system will now be constructed for each image of the board. Recall 
from~(\ref{eqn:amp-segm}) that amplitude-image $B$ typically contains the board, as well as the person holding it.
The very low amplitudes $B_{xy}\approx 0$ of the \emph{black} squares tend 
to be characteristic of the board itself (i.e.\ $B_{xy} \gg 0$ for both the 
white squares and for the rest of $B$). Hence a good estimate of the
board-centre can be obtained by normalizing $B$ to the range $[0,1]$
and then computing a centroid using weights $(1-B_{xy})$.
The centroid, together with the angle $\phi$ from (\ref{eqn:axis}) defines
the Euclidean transformation $\mat{E}$ into local
coordinates, centred on and aligned with the board. 
From now on, unless otherwise stated, it will be assumed that this simple transformation has been performed.


Let $(x_\kappa,y_\kappa,1)^\tp$ be the local coordinates of given point, after
transformation by~$\mat{E}$, with the label $\kappa$ inherited from $\kappa_{xy}$.
Now, by construction, any labelled point is hypothesized to be part of $\mathcal{L}$ 
or $\mathcal{M}$, such that that $\LAMBDA(x_\lambda,y_\lambda,1)^\tp = 0$
or $\MU(x_\mu,y_\mu,1)^\tp = 0$, where
$\LAMBDA$ and $\MU$ are the local coordinates of the relevant lines. These lines can be
expressed as
\begin{equation}
\LAMBDA \simeq (-1,\, \beta_\lambda,\, \alpha_\lambda) 
\mand
\MU \simeq (\beta_\mu,\, -1,\, \alpha_\mu) 
\label{eqn:line-coords}
\end{equation}
with inhomogeneous forms
$x_\lambda = \alpha_\lambda + \beta_\lambda y_\lambda$
and
$y_\mu = \alpha_\mu + \beta_\mu x_\mu$,
such that the slopes 
$|\beta_\kappa| \ll 1$ are \emph{bounded}. In other words, the board is axis-aligned
in local coordinates, and the perspective-induced deviation of any line is less than $45^\circ$.
Furthermore, if the board is visible, then the intercepts $|\alpha_\kappa| \ll \frac{1}{2}(X+Y)$
are bounded in relation to the image size.

Recall from (\ref{eqn:vertices}) that the vertices $\mat{v}_{ij}$ of the board are 
computed as the intersections of $\mathcal{L}$ with $\mathcal{M}$. The resulting points 
can be expressed in the \emph{original} image coordinates as 
$
\mat{v}_{ij} \simeq \mat{E}^{-1} \bigl(\LAMBDA_i \times \MU_j\bigr)^\tp
$, via the inverse transformation.


\subsection{Hough Transform}
\label{sec:hough-transform}

The Hough transform, as used here, maps \emph{points} from the image to \emph{lines}
in the transform. In particular, points along a line are mapped to lines through
a point. This duality between collinearity and concurrency suggests
that a \emph{pencil} of $n$ image-lines will be mapped to a \emph{line} of $n$
transform points, as in fig.~\ref{fig:overview}.

\begin{figure}[!ht]
\centering
\setlength\fboxsep{10pt}
\nofbox{
\begin{minipage}[t]{1\linewidth}
\includegraphics[scale=1]{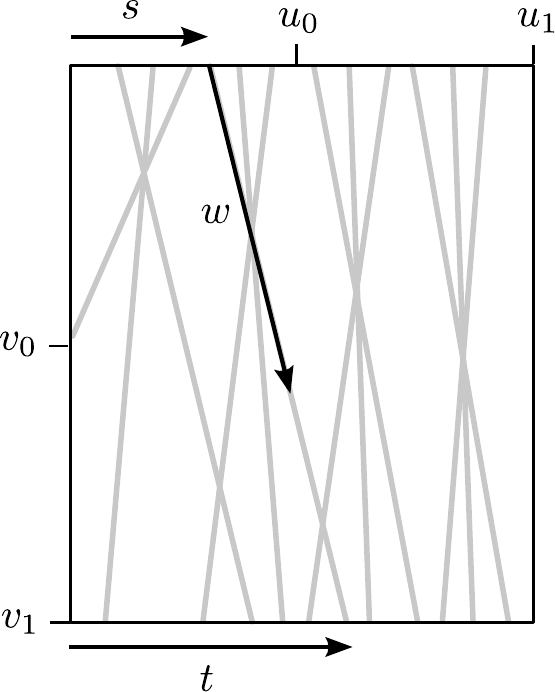} %
\normalsize
\hspace{.25cm}\parbox[b]{0cm}{\vspace*{1cm}\begin{tabbing}
\textbf{for} \=$(x,y)$ \textbf{in} $(0: X) \times (0:Y)$\\[.75ex]
             \>\textbf{if} $\kappa_{xy} \ne \varnothing$\\[.75ex]
             \>\hspace{1.5em}\=$\kappa\leftarrow \kappa_{xy}$\\[.75ex]
             \>\>$s \leftarrow u_\kappa(x,\,y,\,0)$\\[.75ex]
             \>\>$t \leftarrow u_\kappa(x,\,y,\,v_1)$\\[.75ex]
             \>\>$w_1\leftarrow \bigl|(t,v_1) - (s,0)\bigr|$\\[.75ex]
             \>\>\textbf{for} \=$w$ \textbf{in} $\bigl(0:\text{floor}(w_1)\bigr)$\\[.75ex]
             \>\>\>\hspace{-.75em}$H_\kappa \leftarrow H_\kappa \oplus\,
                   \underset{w/w_1}{\mathrm{interp}}\bigl((s,0),\, (t,v_1)\bigr)$\\[-1ex]
             \>\>\textbf{end}\\[-.2ex]
             \>\textbf{endif}\\[-.3ex]
\textbf{end}
\end{tabbing}}
\end{minipage}}
\caption{\textbf{Left:} Hough transform space. \textbf{Right:} Constructing the transform.
Each gradient pixel $(x,y)$ labelled $\kappa \in \{\lambda, \mu\}$ maps to a 
line $u_\kappa(x,y,v)$ in transform $H_\kappa$. The operators 
$H\oplus\mat{p}$ and $\mathrm{interp}_\alpha(\mat{p},\mat{q})$ perform accumulation 
and linear interpolation, respectively. See section \ref{sec:hough-transform} for details.}
\label{fig:transform}
\end{figure}

The transform is implemented as a \twod\ histogram $H(u,v)$, with horizontal and vertical 
coordinates $u\in[0,u_1]$ and $v\in[0,v_1]$. The point $(u_0,v_0)=\frac{1}{2}(u_1,v_1)$ is 
the centre of the transform array. 
Two transforms, $H_\lambda$ and $H_\mu$, will be 
performed, for points labelled $\lambda$ and $\mu$, respectively.
The Hough variables are related to the image coordinates in the following way; let
\begin{gather}
u(x,y,v) = u_0 + x - y (v-v_0), \quad\text{and} \nonumber \\[1ex]
u_\kappa(x,y,v) = 
\begin{cases}
u(x,y,v)
&\text{if } \kappa=\lambda\\[1ex]
u(y,x,v)
&\text{if } \kappa=\mu
\end{cases}
\label{eqn:hough}
\end{gather}
Here $u(x,y,v)$ is the \mbox{$u$-coordinate} of a line (parameterized by~$v$),
which is the Hough-transform of an image-point $(x,y)$.
The Hough intersection point $(u_\kappa^\star,v_\kappa^\star)$ is found by
taking two points $(x,y)$ and $(x',y')$, and 
solving $u_\lambda(x,y,v)=u_\lambda(x',y',v)$, with $x_\lambda$ and
$x_\lambda'$ substituted according to (\ref{eqn:line-coords}).
The same coordinates are obtained by solving $u_\mu(x,y,v)=u_\mu(x',y',v)$, 
and so the result can be expressed as
\begin{equation}
u^\star_\kappa = u_0 + \alpha_\kappa \quad\text{and}\quad
v^\star_\kappa = v_0 + \beta_\kappa
\label{eqn:intersection}
\end{equation}
with labels $\kappa\in\{\lambda,\mu\}$ as usual. 
A peak at $(u_\kappa^\star,v_\kappa^\star)$ evidently maps to a line of intercept 
\mbox{$u_\kappa^\star-u_0$} and slope \mbox{$v_\kappa^\star-v_0$}.
Note that if the perspective 
distortion in the images is small, then $\beta_\kappa\approx 0$, and
all intersection points lie along the horizontal midline $(u,v_0)$ of 
the corresponding transform.
The Hough intersection point $(u_\kappa^\star,v_\kappa^\star)$ can be used
to construct an image-line $\LAMBDA$ or $\MU$,
by combining (\ref{eqn:intersection}) with (\ref{eqn:line-coords}),
resulting in
\begin{equation}
\begin{aligned}
\LAMBDA &\leftarrow
\bigl(-1,\
v^\star_\lambda - v_0,\
u^\star_\lambda - u_0\bigr) \\[1ex]
\MU &\leftarrow
\bigl(v^\star_\mu-v_0,\
-1,\
u^\star_\mu - u_0 \bigr).
\label{eqn:line-coords-hough}
\end{aligned}
\end{equation}
The two Hough transforms are computed by the procedure in fig.~\ref{fig:transform}. 
Let $H_\kappa$ refer to $H_\lambda$ or $H_\mu$, according to the label $\kappa$ of the
 point $(x,y)$. For each accepted point, the corresponding 
line (\ref{eqn:hough}) intersects the top and bottom of the $(u,v)$ array at points 
$(s,0)$ and $(t,v_1)$ respectively. The resulting segment, of length $w_1$,
is evenly sampled, and $H_\kappa$ is incremented at each of the constituent points.

The procedure in fig.~\ref{fig:transform} makes use of the following functions.
Firstly, $\mathrm{interp}_\alpha(\mat{p}, \mat{q})$, with $\alpha\in[0,1]$, returns the 
convex combination $(1-\alpha)\mat{p}+\alpha\mat{q}$. Secondly, the `accumulation' 
$H \oplus (u,v)$ is equal to $H(u,v) \leftarrow H(u,v) + 1$ if $u$ and $v$ are integers. 
In the general case, however, the four pixels closest to $(u,v)$ are updated by the
corresponding bilinear-interpolation weights (which sum to one). This weighting-scheme, combined with the large number of gradient-vectors that are processed, tends to produce a relatively smooth histogram.

\subsection{Hough Analysis}
\label{sec:hough-analysis}

The local coordinates defined in sec.~\ref{sec:local-coordinates} ensure that the
two Hough transforms $H_\lambda$ and $H_\mu$ have the same characteristic structure.
Hence the subscripts $\lambda$ and $\mu$ will be suppressed for the moment. Recall that 
each Hough cluster corresponds to a line in the image space, and that a collinear set of 
Hough clusters corresponds to a pencil of lines in the image space, as in fig~\ref{fig:overview}. 
It follows that all lines in a pencil can be detected simultaneously, by \emph{sweeping} the 
Hough space $H$ with a line that cuts a \oned\ slice through the histogram.

\begin{figure}[t!]
\centering
\begin{large}
\setlength\fboxsep{10pt}
\nofbox{
\begin{minipage}[t]{1\linewidth}
\includegraphics[scale=1]{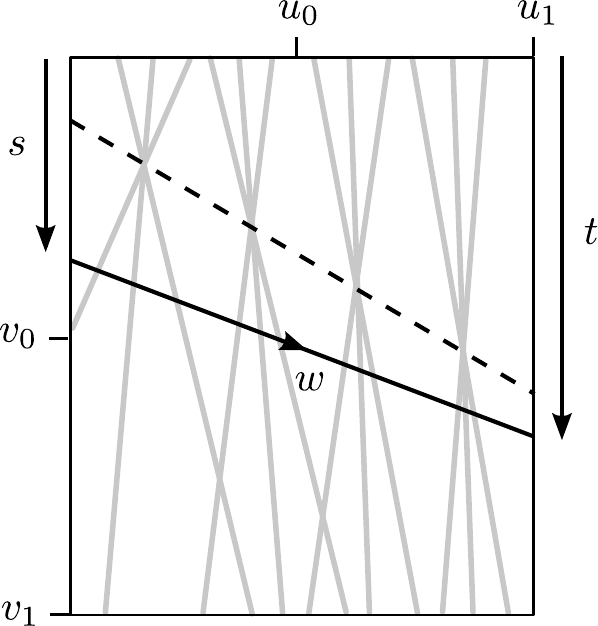} %
\hspace*{.5cm}\normalsize
\parbox[b]{0cm}{\vspace*{.5cm}\begin{tabbing}
\textbf{for} \=$(s,t)$ \textbf{in} $(0 : v_1) \times (0 : v_1)$\\[.75ex]
             \>$w_1 = \bigl|(u_1,t) - (0,s)\bigr|$\\[.75ex]
             \>\textbf{for} \=$w$ \textbf{in} $\bigl(0 : \text{floor}(w_1)\bigr)$\\[1ex]
             \>\>$(u,v) \leftarrow \underset{w/w_1}{\mathrm{interp}}\bigl((0,s),\, (u_1,t)\bigr)$\\[1ex]
             \>\>$h_\lambda^{st}(w) \leftarrow H_\lambda(u,v)$\\[1.25ex]
             \>\>$h_\mu^{st}(w) \leftarrow H_\mu(u,v)$\\[.75ex]
             \>\textbf{end}\\
\textbf{end}
\end{tabbing}}
\end{minipage}}
\end{large}
\caption{\textbf{Left:} Hough transform space. \textbf{Right:} Searching the transform. A line $h^{st}_\kappa(w)$, with end-points $(0,s)$ and $(u_1,t)$, 
is swept through each Hough transform $H_\kappa$. A total of $v_1\times v_1$ \oned\ histograms
$h^{st}_\kappa(w)$ are computed in this way.
See section \ref{sec:hough-analysis} for details.}
\label{fig:analysis}
\end{figure}

Recall from section \ref{sec:hough-transform} that the Hough peaks are most 
likely to lie along a horizontal axis (corresponding to a fronto-parallel pose of the
board). Hence a suitable parameterization of the sweep-line is to vary one endpoint 
$(0,s)$ along the left edge, while varying the other endpoint $(u_1,t)$ along the 
right edge, as in fig.~\ref{fig:analysis}. This scheme has the desirable property
of sampling more densely around 
the midline $(u,v_0)$. It is also useful to note that the sweep-line parameters $s$ and $t$
can be used to represent the apex of the corresponding pencil. The local
coordinates $\mat{p}$ and $\mat{q}$ are
$
\mat{p} \simeq (\LAMBDA_s \times \LAMBDA_t)^\tp
$
and
$
\mat{q} \simeq (\MU_s \times \MU_t)^\tp
$
where $\LAMBDA_s$ and $\LAMBDA_t$ are obtained 
from (\ref{eqn:line-coords}) by setting 
$(u_\lambda^\star,v_\lambda^\star)$ to $(0,s)$ and $(u_1,t)$
respectively, and similarly for $\MU_s$ and $\MU_t$.


The procedure shown in fig.~\ref{fig:analysis} is used to analyze the
Hough transform. The sweep-line with parameters $s$ and $t$ has the form of a \oned\ 
histogram $h_\kappa^{st}(w)$. The integer index \mbox{$w\in(0:w_1)$} is equal to the 
Euclidean distance $|(u,v) - (0,s)|$ along the sweep-line. 
The procedure shown in fig.~\ref{fig:analysis} makes further use of the
interpolation operator that was defined in section 
\ref{sec:hough-transform}.
Each sweep-line $h_\kappa^{st}(w)$, constructed by the above process, will contain 
a number of isolated clusters: \mbox{$\bigl|\mathrm{clusters}(h_\kappa^{st})\bigr| \ge 1$}. These  clusters are simply defined as runs of non-zero values in $h_\kappa^{st}(w)$.
The existence of separating zeros is, in practice, highly reliable 
when the sweep-line is close to the true solution. This is simply because the
Hough data was thresholded in (\ref{eqn:labelling}), and strong gradients are not
found \emph{inside} the chessboard squares. The representation of the clusters, 
and subsequent evaluation of each sweep-line, will now be described.

The label $\kappa$ and endpoint parameters $s$ and $t$ will be suppressed,
in the following analysis of a single sweep-line, for clarity. Hence let 
$w \in (a_c:b_c)$ be the interval that contains the  \mbox{$c$-th} cluster in $h(w)$.
The score and location of this cluster are defined as the mean value and
centroid, respectively:
\begin{align}
\underset{c}{\text{score}}(h)
= \frac{\sum_{w = {a_c}}^{b_c} h(w)}{1+b_c-a_c}\\[1.5ex]
w_c
= a_c + \frac{\sum_{w = {a_c}}^{b_c} h(w)\mspace{2mu} w}
{\sum_{w = {a_c}}^{b_c} h(w)}
\end{align}
More sophisticated definitions are possible, based on quadratic interpolation around
each peak. However, the mean and centroid give similar results in practice.
A total score must now be assigned to the sweep-line, based on the 
scores of the constituent clusters. If $n$ peaks are sought, then the total score
is the sum of the highest $n$ cluster-scores. But if there are fewer than $n$ 
clusters in $h(w)$, then this cannot be a solution, and the score is zero:
\begin{equation}
{\Sigma}^{\mspace{1mu}n}(h) =
\begin{cases}
\sum_{i=1}^n \underset{c(i)}{\mathrm{score}}(h)
&\text{if } n \le \bigl|\mathrm{clusters}(h)\bigr| \\[.75ex]
0 &\text{otherwise}
\end{cases}
\label{eqn:total-score}
\end{equation}
where $c(i)$ is the index of the \mbox{$i$-th} highest-scoring cluster.
The optimal clusters are those in the sweep-line that maximizes~(\ref{eqn:total-score}).
Now, restoring the full notation, the score of the optimal sweep-line in the transform
$H_\kappa$ is
\begin{equation}
{\Sigma}_\kappa^{\mspace{1mu}n}
\leftarrow \max_{s,\,t} \,
\underset{n}{\mathrm{score}}
\bigl(h_\kappa^{st}\bigr).
\end{equation}
One problem remains: it is not known in advance whether there should 
be $\ell$ peaks in $H_\lambda$ and $m$ in $H_\mu$, or vice versa.
Hence all four combinations, $\Sigma^\ell_\lambda$, $\Sigma^m_\mu$, $\Sigma^\ell_\mu$, $\Sigma^m_\lambda$ 
are computed. The ambiguity between pencils $(\mathcal{L}, \mathcal{M})$ 
and labels $(\lambda,\mu)$ can then be resolved, by picking the solution
with the highest \emph{total} score:
\begin{equation}
\bigl(
\mathcal{L}, \mathcal{M}
\bigr) \Leftrightarrow
\begin{cases}
(\lambda,\mu) &\text{if } 
{\Sigma}^\ell_\lambda + {\Sigma}^m_\mu > 
{\Sigma}^\ell_\mu + {\Sigma}^m_\lambda\\
(\mu,\lambda) &\text{otherwise}.
\end{cases}
\label{eqn:correspondence}
\end{equation}
Here, for example, 
$
\bigl(
\mathcal{L}, \mathcal{M}
\bigr) \Leftrightarrow
(\lambda,\mu)
$
means that there is a pencil of $\ell$ lines in $H_\lambda$ and a pencil of 
$m$ lines in $H_\mu$. The procedure in (\ref{eqn:correspondence}) is based on
the fact that the complete solution must consist of $\ell + m$ clusters.
Suppose, for example, that there are $\ell$ good clusters in $H_\lambda$, and 
$m$ good clusters in $H_\mu$. Of course there are also $\ell$ good clusters
in $H_\mu$, because $\ell < m$ by definition.
However, if only $\ell$ clusters are taken from $H_\mu$, then an additional
$m-\ell$ weak or non-existent clusters must be found in $H_\lambda$, and so
the total score ${\Sigma}^\ell_\mu + {\Sigma}^m_\lambda$ would not be maximal.

It is straightforward, for each centroid $w_c$ in the optimal sweep-line $h_\kappa^{st}$,
to compute the \twod\ Hough coordinates
\begin{equation}
\bigl(u^\star_\kappa,\,v^\star_\kappa\bigr) \leftarrow
\underset{w_c/w_1}{\mathrm{interp}}\bigl((0,s),\,(u_1,t)\bigr)
\end{equation}
where $w_1$ is the length of the sweep-line, as in fig.~\ref{fig:analysis}. 
Each of the resulting $\ell m$ points are mapped to image-lines, according to 
(\ref{eqn:line-coords-hough}).
The vertices $\mat{v}_{ij}$ are then computed from (\ref{eqn:vertices}).
The order of 
intersections along each line is preserved by the Hough transform, and so the $ij$ 
indexing is automatically consistent. 

Finally, in order to minimize any effects of discretization, it would be possible to perform quadratic (or other) interpolation around the maximal sweep-line score, with respect to the end-point parameters. In practice, owing to the high resolution of the sweep-procedure around zero-slope (as described above), this does not prove to be necessary.




\subsection{Decision Functions}
\label{sec:decision-functions}

The analysis of section \ref{sec:hough-analysis} returns estimated pencils
$(\mathcal{L},\mathcal{M})$, even if the board is not visible in 
the image. Hence it is necessary to evaluate the quality of the solution.
One possibility would be to test a statistic based on the scores 
in (\ref{eqn:correspondence}).
However, a more robust approach is to test whether the solution 
$(\mathcal{L},\mathcal{M})$ satisfies certain geometric constraints
that were \emph{not} used in the estimation process.
 
There are, in practice, two types of mis-detections. In the 
first case a minority of the lines are not aligned with the chessboard 
pattern, and so the solution is \textbf{corrupted}.
In the second case the lines include an external edge of the 
chessboard pattern, and so the entire solution is \textbf{displaced}.
These two types of error, which may co-occur, are addressed below.

Recall that vertex $\mat{v}_{ij}$ is the intersection between lines 
$\LAMBDA_i$ and $\MU_j$, as in (\ref{eqn:vertices}).
Corrupted estimates can be detected by noting that \emph{cross ratios}
of distances between quadruples of vertices should be near unity, given 
that the observations are projectively related to a regular grid~\cite{hartley-2000,ha-2009}.
In practice,
extremely foreshortened boards cannot be detected (owing to limited
resolution), and so it suffices to use an \emph{affine} test. 
In particular, for line $\LAMBDA_i$, consider the ratio of the 
vertex-intervals starting at positions $j$ and $k$:
\begin{equation}
F_{jk}(\LAMBDA_i) = 
\frac
{|\mat{v}_{i(j+1)} - \mat{v}_{ij}|}
{|\mat{v}_{i(k+1)} - \mat{v}_{ik}|}.
\end{equation}
These ratios are tested along the first and last line in each 
pencil, and so a suitable set of decision functions is
\begin{gather}
\bigl|1 - F_{jk}(\LAMBDA_i)\bigr| \le f
\quad\text{for all } i,j,k\ \label{eqn:test-corrupted}\\[.75ex]
i = \{1,\ell\},
\quad j,k \in (1 : m-1), \,\,\, j\ne k
\nonumber
\end{gather}
for a small positive threshold $f$. The analogous tests are applied
to the other pencil, $\mathcal{M}$. If any of the tests 
(\ref{eqn:test-corrupted}) are failed, then the estimate
$(\mathcal{L},\mathcal{M})$ is rejected.

Displaced estimates can be determined as follows. 
Let $\mat{w}\in \LAMBDA_i$ be an image-point on the line segment 
between vertices $\mat{v}_{i1}$ and $\mat{v}_{im}$ that lies `inside' the pencil 
$\mathcal{M}$. If this segment is strictly inside the chessboard pattern, then 
there should be equal numbers of black-white and white-black transitions across it.
If, on the other hand, the segment is along the perimeter of the pattern, 
then the two types of transition will be very imbalanced, as the black
squares are `missing' on one side of the perimeter.
Let $(\xi_\mat{w},\eta_\mat{w})$ be the gradient at edge-point $\mat{w}$, and let
$\LAMBDA^{\!\star}=(\alpha,\beta,\gamma)$ be the normalized line 
coordinates, such that $\alpha^2+\beta^2 = 1$. Now the dot-product 
$\LAMBDA^{\!\star} (\xi_\mat{w},\eta_\mat{w},0)^\tp$ is the projection 
of the gradient onto the edge-normal at $\mat{w}$, and so the amount of black-white 
vs.\ white-black transitions can be measured by the ratio of the summed 
positive and negative projections,
\begin{gather}
G(\LAMBDA_i) = 
\frac
{\sum_{k=0}^{k\le d}\, \mathrm{pos}\bigl(\LAMBDA_i^{\!\star} \, (\xi_\mat{w},\eta_\mat{w},0)^\tp\bigr)}
{\sum_{k=0}^{k\le d}\, \mathrm{neg}\bigl(\LAMBDA_i^{\!\star} \, (\xi_\mat{w},\eta_\mat{w},0)^\tp\bigr)}
\quad\text{where}\\[1ex]
d = |\mat{v}_{im} - \mat{v}_{i1}| \quad\text{and}\quad
\mat{w} = \underset{k/d}{\mathrm{interp}}(\mat{v}_{i1},\, \mat{v}_{im}).
\end{gather}
Here the function $\mathrm{pos}(x)$ returns $x$ if $x>0$, or zero otherwise, and 
$\mathrm{neg}(x)$ returns $|x|$ if $x<0$, or zero otherwise.
The ratio $G(\LAMBDA)$ should be close to unity for all internal segments, 
and so an appropriate set of decision functions is
\begin{equation}
\bigl|1 - G(\LAMBDA_i\bigr)\bigr| \le g \quad\text{for all } i = 1:\ell
\label{eqn:test-displaced}
\end{equation}
where $g$ is a small positive threshold. The same test is applied to 
the $j=(1:m)$ segments between $\mat{v}_{1j}$ and $\mat{v}_{\ell j}$,
in pencil~$\mathcal{M}$.

The thresholds $f$ and $g$ in tests (\ref{eqn:test-corrupted}) and (\ref{eqn:test-displaced}) were fixed once and for all in the experiments, 
such that no false-positive detections (which are unacceptable for calibration purposes) 
were made, across \emph{all} data-sets. Furthermore, the same thresholds were used for all cameras.


\section{Results}
\label{sec:results}


The new method is evaluated in two ways. 
Firstly, in section \ref{sec:robustness}, the robustness of the gradient-labelling method to perspective distortion is investigated.
Secondly, in section \ref{sec:comparison} the overall detection performance is compared, over a data-set of several hundred real images, to the most commonly-used (OpenCV) detection method. 

It is important, at the outset, to clarify the nature of the evaluation. 
The ultimate objective of a board-detection algorithm, for calibration purposes, is to detect as many boards as possible, in the greatest variety of poses. In particular, \emph{good coverage of the entire 3D scene-volume}, by the board vertices, is required for subsequent extrinsic calibration procedures. However, it is also essential that the vertices be \emph{accurate}, in the sense of geometric error; in particular, there must be no `false' detections or mis-labellings of the vertices. In practice, this means that the parameters of any detection algorithm must be set conservatively. Having done this, then any additional detection is beneficial for subsequent extrinsic calibration, provided that it does not increase the overall geometric error. The evaluation in \ref{sec:comparison} is based on large real-world calibration sets, which inevitably contain many problematic images (including those in which the board is not fully visible to one or more cameras). In order to make a real evaluation, there was no attempt to avoid or subsequently remove these images from the data-sets.
 

\subsection{Geometric robustness}
\label{sec:robustness}
This section investigates whether the method can reliably identify two pencils of lines, in the presence of perspective distortion. In particular, it must be shown that the method does not require the midlines of the two pencils to be orthogonal. Ideally, the evaluation would be performed on a controlled image-set, showing a board at all possible orientations. In practice, it is very hard to obtain such an image-set, without using a mechanical mounting for the board. However, it is possible to accurately simulate the \emph{geometric} effects of foreshortening, by applying a 2D homography to real image-data. This makes it straightforward to construct a uniform distribution of board orientations, including all slants $\varphi$. Furthermore, each classified pixel, labelled $\kappa_\varphi$, is \emph{in correspondence} with the fronto-parallel `best case', labelled $\kappa_0$. The latter can therefore be used as ground-truth measurements for the evaluation, as explained below. This procedure, importantly, allows us to isolate \emph{geometric} effects from \emph{sampling} effects, which leads to a better understanding.

In more detail, a single fronto-parallel (zero slant) image is processed as described in section~\ref{sec:preprocessing}, yielding a gradient vector $(\xi_0,\eta_0,1)$, and a label $\kappa_0$ at each pixel. A rotation matrix $\mat{R}_\varphi = \mat{R}\bigl(\varphi,\mat{w}(\vartheta)\bigr)\,\mat{R}(\omega_,\mat{z})$ is then constructed, using angle-axis factors $\mat{R}(\cdot,\cdot)$. The rotation has the effect of \emph{slanting} the board by an angle $\varphi$, around an axis $\mat{w}(\vartheta)$. This axis is perpendicular to the optical axis $\mat{z}$, with $\vartheta$ being the \emph{tilt} angle. The second factor, $\mat{R}(\omega,\mat{z})$, is a cyclo-rotation around the optical axis $\mat{z}$.
The important variable is the slant angle, $\varphi$, which is systematically varied from zero (fronto-parallel) to $90^\circ$. At each slant, the tilt and cyclorotation angles $\vartheta$ and $\omega$ are sampled 100 times from the uniform distribution on $[0,2\pi]$.

\begin{figure}[ht!]
\centering
\includegraphics[scale=0.5]{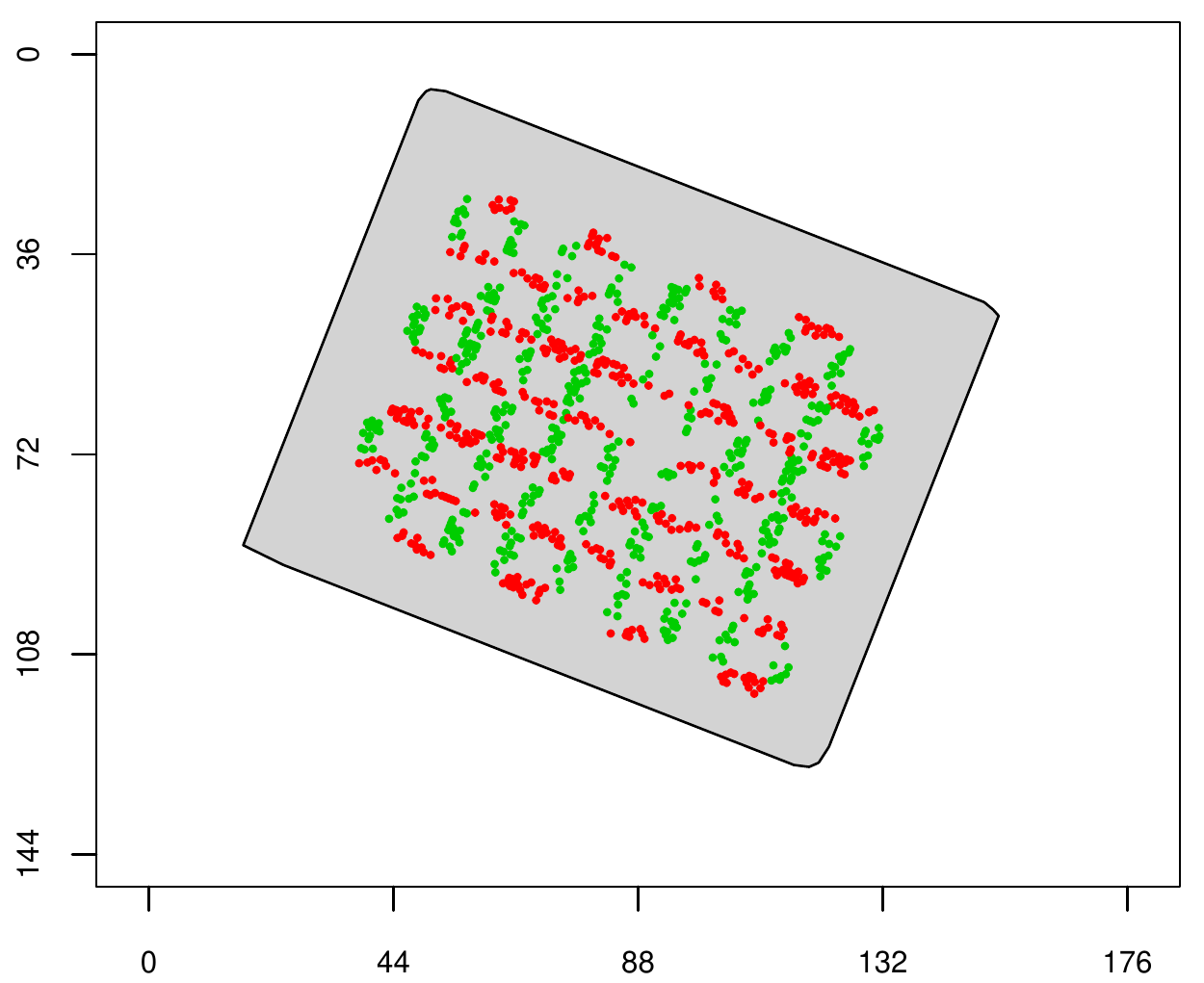}\\
\includegraphics[scale=0.5]{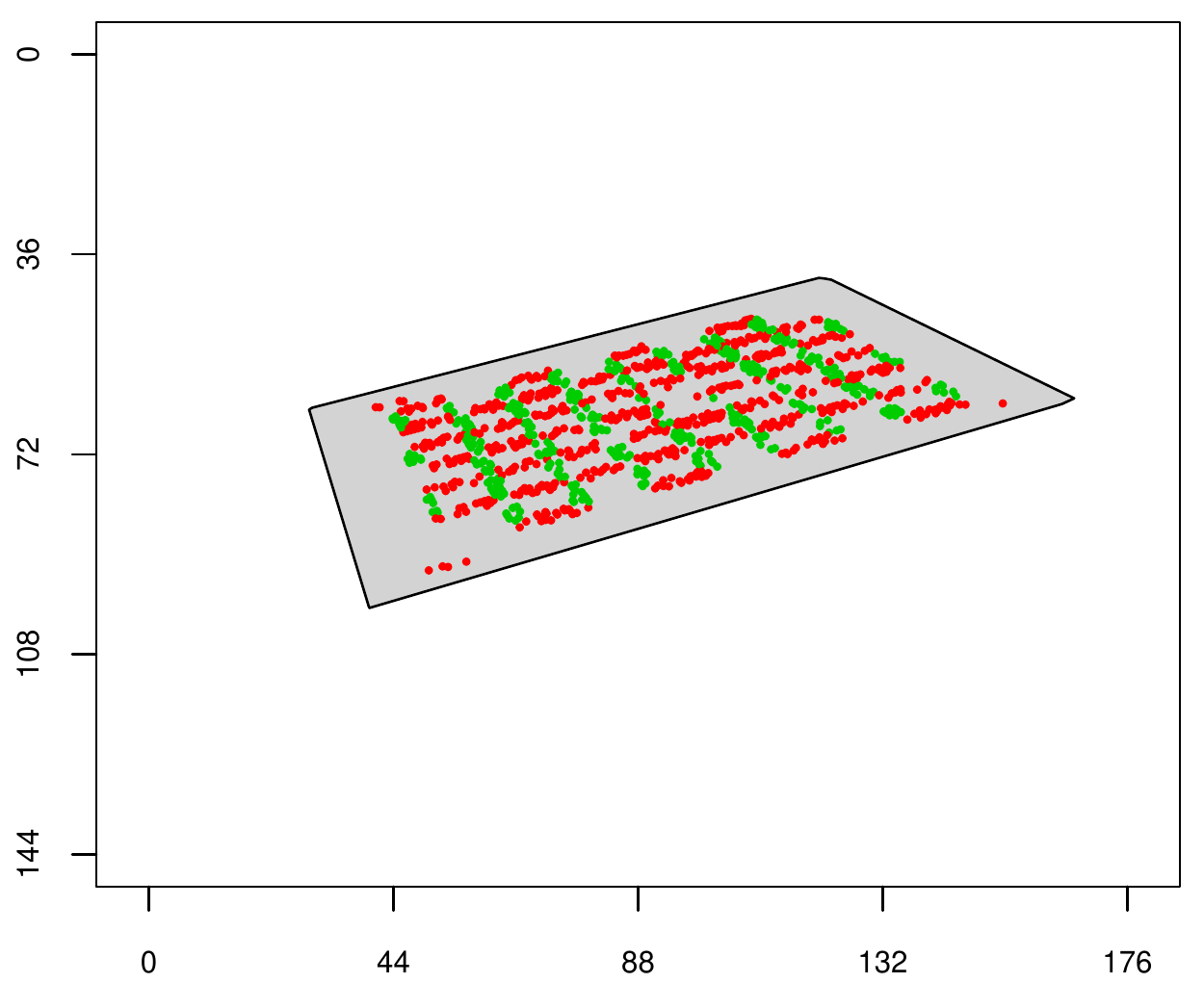}
\caption{Example trials from the slant experiment. Top: A cyclo-rotation of the original gradient-vectors, which are easily classified (except for saturated highlight in the centre), as there is zero slant $\varphi$. Bottom: the classification is more difficult at a slant of $\varphi=70^\circ$. A total of 5000 gradient vectors were sampled, but only the labelled pixels are shown. The grey rectangles are the convex hulls of the 5000 samples. A total of 1200 trials were performed; the results are shown in fig.~\ref{fig:slant-plot}.}
\label{fig:slant-boards}
\end{figure}

A homography $\mat{H}_\varphi$ is now constructed from the rotation matrix $\mat{R}_\varphi$, and applied to the gradient-vectors, so that $(\xi_\varphi,\eta_\varphi,1) = (\xi_0,\eta_0,1)\mat{H}_\varphi^{-1}$. The transformed gradients are then re-classified, and the `slanted' label $\kappa_\varphi$ is compared to the `fronto-parallel' label $\kappa_0$, at each pixel. 
This can be visualized as a comparison between the two classifications in fig.~\ref{fig:slant-boards}.
Consistency of the slanted classification is defined as the proportion of unchanged labels, over all edges. If $\delta(\kappa_\varphi, \kappa_0)$ is the indicator function that is~1 if $\kappa_\varphi = \kappa_0$, and~0 otherwise, then
\[
\text{consistency} = 
\frac
{\sum_{\{\kappa_0 \ne \varnothing\}} \delta(\kappa_\varphi, \kappa_0)}
{\bigl|\{\kappa_0 \ne \varnothing\}\bigr|}
\]
where $\{\kappa_0 \ne \varnothing\}$ is the set of edge-pixels (i.e.~large gradients), and $\bigl|\{\kappa_0 \ne \varnothing\}\bigr|$ is the number of pixels in this set. This statistic is relatively strict, in that it ignores the easy task of classifying the non-edge pixels, for which $\kappa=\varnothing$.

\begin{figure}[ht!]
\centering
\includegraphics[scale=0.65]{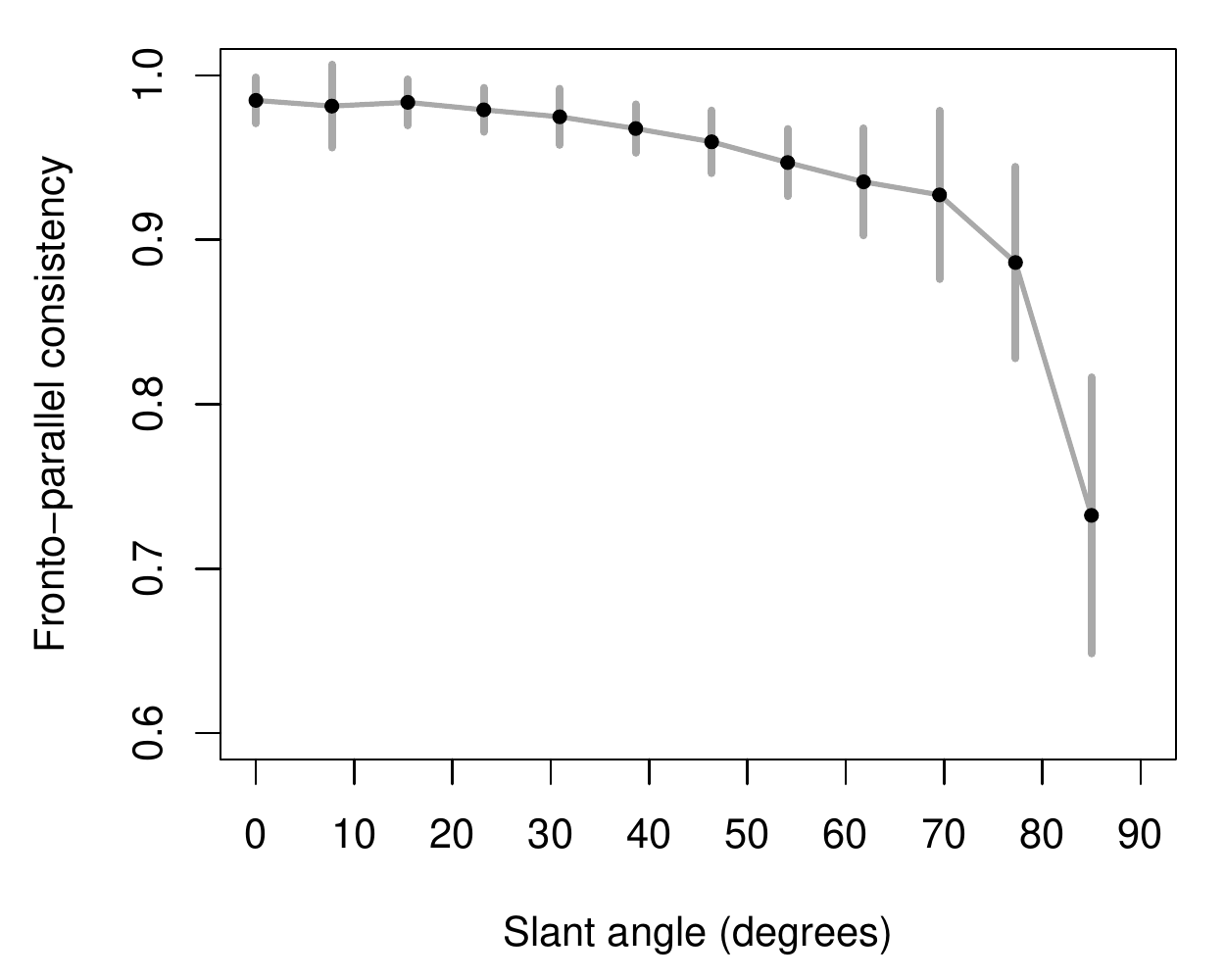}
\caption{Edge-classification accuracy, over 1200 trials, as a function of slant-angle. The labelling becomes less accurate as the foreshortening increases, and eventually becomes undefined at $90^\circ$ (when only the edge of the board can be seen). The labelling is robust to slants of more than $70^\circ$. The error bars show $\pm 1$ standard deviation per 100 trials; the variation is due to the random tilt-angle on each trial.}
\label{fig:slant-plot}
\end{figure}

The results of the experiment are shown in fig.~\ref{fig:slant-plot}, from which it can be seen that the labelling is robust to slants of more than $70^\circ$. This finding validates the basic detection principle, in relation to the geometry of perspective projection. However, in this experiment, the low-level image-processing was performed in the original fronto-parallel images. This was to avoid confounding geometric effects with sampling effects. A complete evaluation, using a real calibration data-set, is performed in the following section.


\subsection{Comparison with heuristic methods}
\label{sec:comparison}

The method was formally tested on five multi-view data-sets, captured by
Mesa Imaging Swiss-Ranger SR-4000 \textsc{\tof{}} cameras \cite{mesa-2010}.
We used between two and four cameras per data-set (see table~\ref{tab:results}), all performing simultaneous capture.
The sizes $u_1$ and $v_1$ of the Hough array were set to 1.5 times the average dimension, $(X+Y)/2$, of the input images.

We compare our results to the widely-used OpenCV detector \cite{bradski-2008}. Both the OpenCV and Hough detections were refined
by the OpenCV subpixel routine, which adjusts the given point to minimize
the discrepancy with the image-gradient around the chequerboard corner
\cite{bradski-2008,datta-2009}. Specifically, if the true vertex has coordinates 
$(x_0,y_0)$, and $(\xi_{xy}, \eta_{xy})$ is the image-gradient at a nearby point
$(x,y)$, then
\begin{equation}
(\xi_{xy},\, \eta_{xy}) \cdot (x-x_0,\, y-y_0) \approx 0.
\label{eqn:photometric}
\end{equation}
This is because if the magnitude of $(\xi_{xy}, \eta_{xy})$ is not negligible, then $(x,y)$ 
must be \emph{on} a black-white edge. Hence $(\xi_{xy}, \eta_{xy})$ must be perpendicular 
to the edge, while $(x-x_0, y-y_0)$ must be parallel to it. 

Table~\ref{tab:results} shows the number of 
true-positive detections by each method, as well as the number of detections
common to both methods. The geometric error
is the discrepancy from the `ideal' board, after fitting the latter by
the optimal homography (initialized by the DLT method, and refined by Levenberg-Marquardt optimization \cite{hartley-2000}). This is by far the
most useful measure, as it is directly related to the role of the detected
vertices in subsequent calibration algorithms (e.g.~bundle-adjustment \cite{hartley-2000}), and also has a simple interpretation in pixel-units.
The photometric error is the RMS gradient residual (\ref{eqn:photometric}) computed over
a $5\times 5$ window. This measure is worth considering,
because it is the criterion minimized by the subpixel optimization, but it is much
less useful than the geometric error.

\begin{table*}[!ht]
\centering
\begin{small}
\begin{tabular}{|c|ccc|cc|cc|}
\hline
 & \multicolumn{3}{|c|}{{\bf Number detected}} & \multicolumn{2}{|c|}{{\bf Geometric error}} & \multicolumn{2}{|c|}{Photometric error}\\ \hline
Set / Camera & OCV & HT & Both & OCV & HT & OCV & HT\\ \hline\hline
1 / 1 &  19 & 34 & 13  & 0.2263  & 0.1506  & 0.0610  & 0.0782  \\
1 / 2 &  22 & 34 & 14  & 0.1819  & 0.1448  & 0.0294  & 0.0360  \\
1 / 3 &  46 & 33 & 20  & 0.1016  & 0.0968  & 0.0578  & 0.0695  \\
1 / 4 &  26 & 42 & 20  & 0.2044  & 0.1593  & 0.0583  & 0.0705  \\ \hline
2 / 1 &  15 & 27 & 09  & 0.0681  & 0.0800  & 0.0422  & 0.0372  \\
2 / 2 &  26 & 21 & 16  & 0.0939  & 0.0979  & 0.0579  & 0.0523  \\
2 / 3 &  25 & 37 & 20  & 0.0874  & 0.0882  & 0.0271  & 0.0254  \\ \hline
3 / 1 &  14 & 26 & 11  & 0.1003  & 0.0983  & 0.0525  & 0.0956  \\
3 / 2 &  10 & 38 & 10  & 0.0832  & 0.1011  & 0.0952  & 0.1057  \\
3 / 3 &  25 & 41 & 21  & 0.1345  & 0.1366  & 0.0569  & 0.0454  \\
3 / 4 &  18 & 23 & 10  & 0.1071  & 0.1053  & 0.0532  & 0.0656  \\ \hline
4 / 1 &  16 & 21 & 14  & 0.0841  & 0.0874  & 0.0458  & 0.0526  \\
4 / 2 &  45 & 53 & 29  & 0.0748  & 0.0750  & 0.0729  & 0.0743  \\
4 / 3 &  26 & 42 & 15  & 0.0954  & 0.0988  & 0.0528  & 0.0918  \\ \hline
5 / 1 &  25 & 37 & 18  & 0.0903  & 0.0876  & 0.0391  & 0.0567  \\
5 / 2 &  20 & 20 & 08  & 0.2125  & 0.1666  & 0.0472  & 0.0759  \\
5 / 3 &  39 & 36 & 24  & 0.0699  & 0.0771  & 0.0713  & 0.0785  \\
5 / 4 &  34 & 35 & 19  & 0.1057  & 0.1015  & 0.0519  & 0.0528  \\ \hline
6 / 1 &  29 & 36 & 20  & 0.1130  & 0.1203  & 0.0421  & 0.0472  \\
6 / 2 &  35 & 60 & 26  & 0.0798  & 0.0803  & 0.0785  & 0.1067  \\
\hline\hline
Mean: & {\bf 25.75} & {\bf 34.8} & 16.85 & {\bf 0.1157}  & {\bf 0.1077} & 0.0547 & 0.0659 \\ \hline
\end{tabular}
\end{small}
\vspace{2ex}
\caption{Results over six multi-\textsc{\tof{}} camera-setups.
Total detections for the OpenCV ($N=515$) vs.\ Hough Transform ($N=696$) method
are shown, as well as the accuracy of the estimates. 
Geometric error is in pixels. The chief
conclusion is that the HT method detects 35\% more boards, 
and slightly reduces the average geometric error.
The increased number and variety of detected boards is very beneficial for extrinsic calibration procedures, which require the input-points to be spread throughout the entire scene-volume.
}
\label{tab:results}
\end{table*}

The proposed Hough-based method detects 35\% more boards than the OpenCV
method, on average. There is also a slight reduction in average
geometric error, even though the additional boards were more problematic to
detect. 
Even if OpenCV had detected these boards, it is likely that their inclusion would have increased the geometric error even further.
It may also be noted that the new method is \emph{dramatically} better in five cases, in the sense that it detects an additional ten or more boards, while also \emph{reducing} the geometric error.
These results should not be surprising, because the new method uses a
very strong model of the global board-geometry.



\begin{figure}[t!]
\centering
\includegraphics[width=.49\linewidth]{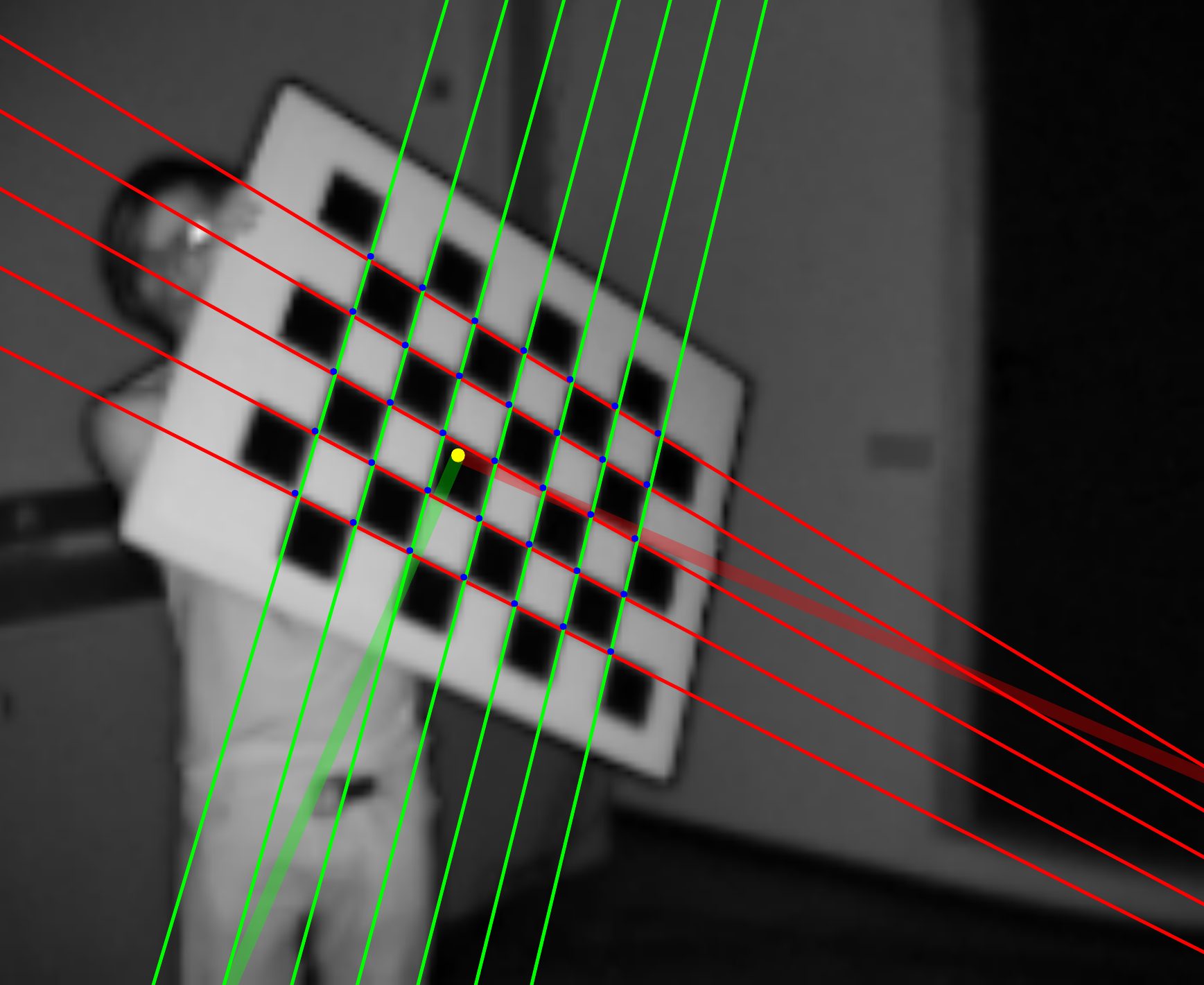}
\includegraphics[width=.49\linewidth]{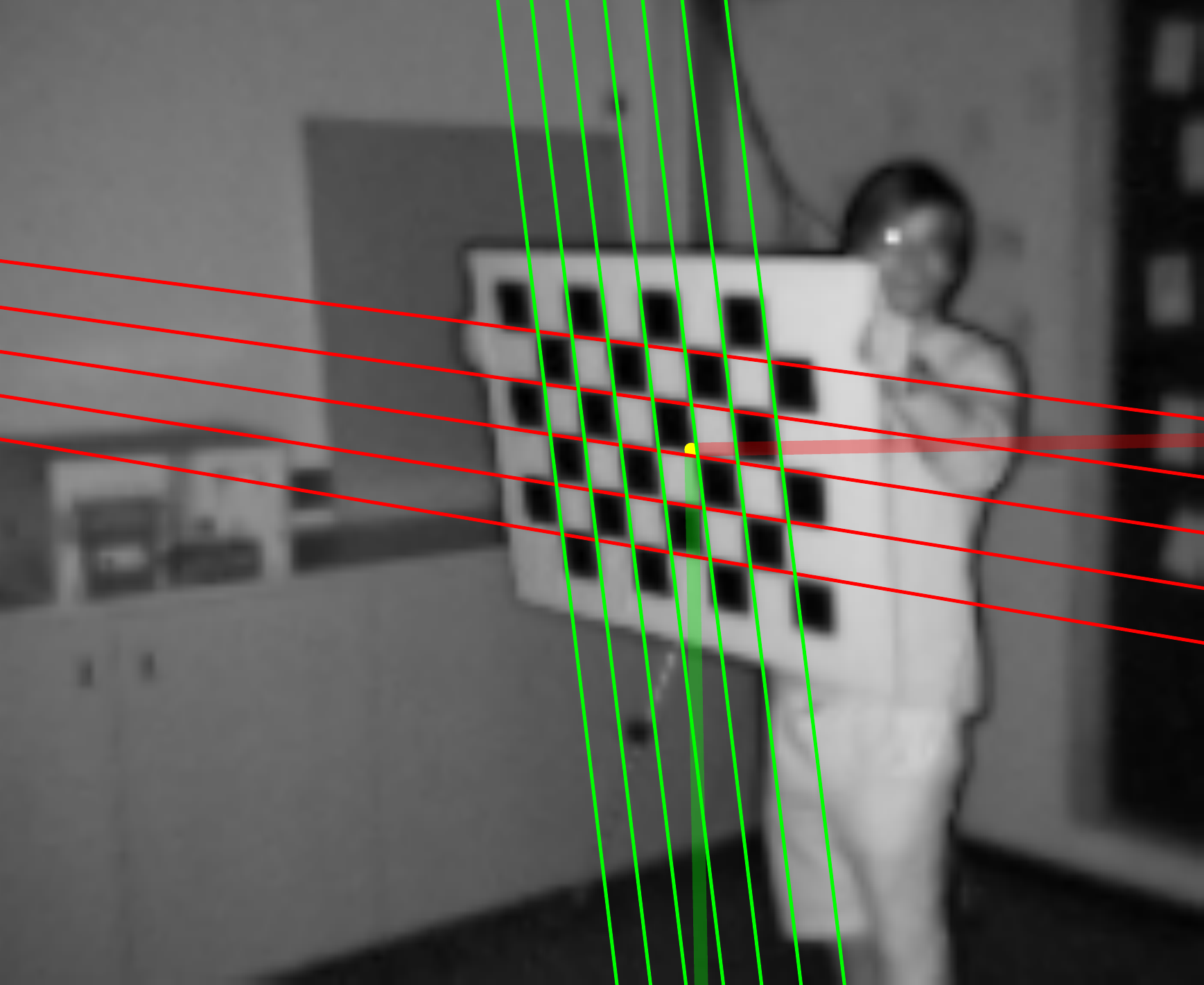}
\caption{Example detections in $176\times 144$ \textsc{\tof{}} amplitude images. 
The yellow dot (one-pixel radius) is the estimated centroid of the
board, and the attached thick translucent lines are the estimated 
axes. The board on the right, which is relatively distant and slanted,
was not detected by OpenCV.}
\label{fig:boards-1}
\end{figure}

\begin{figure}[ht!]
\centering
\includegraphics[width=.49\linewidth]{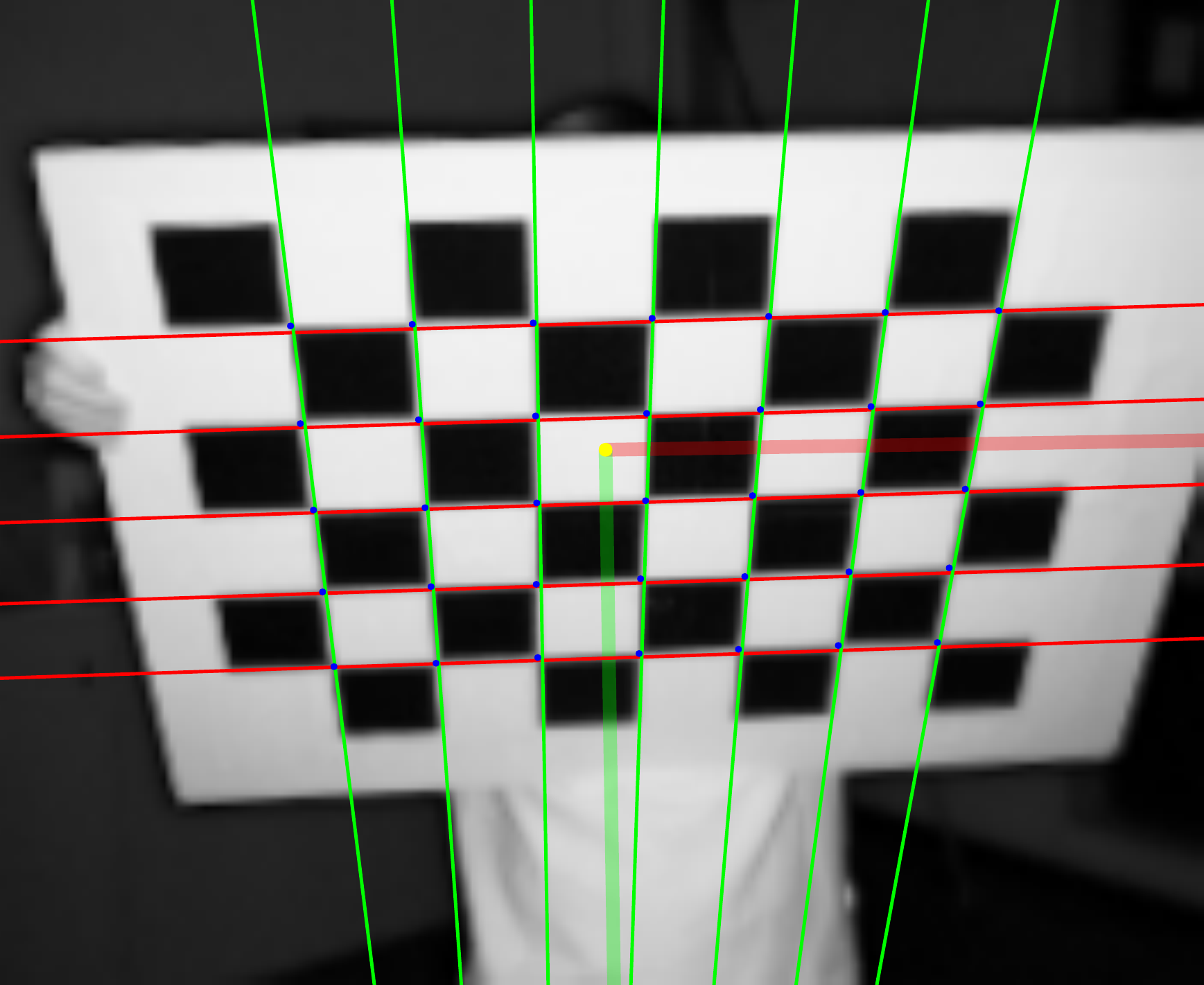}
\includegraphics[width=.49\linewidth]{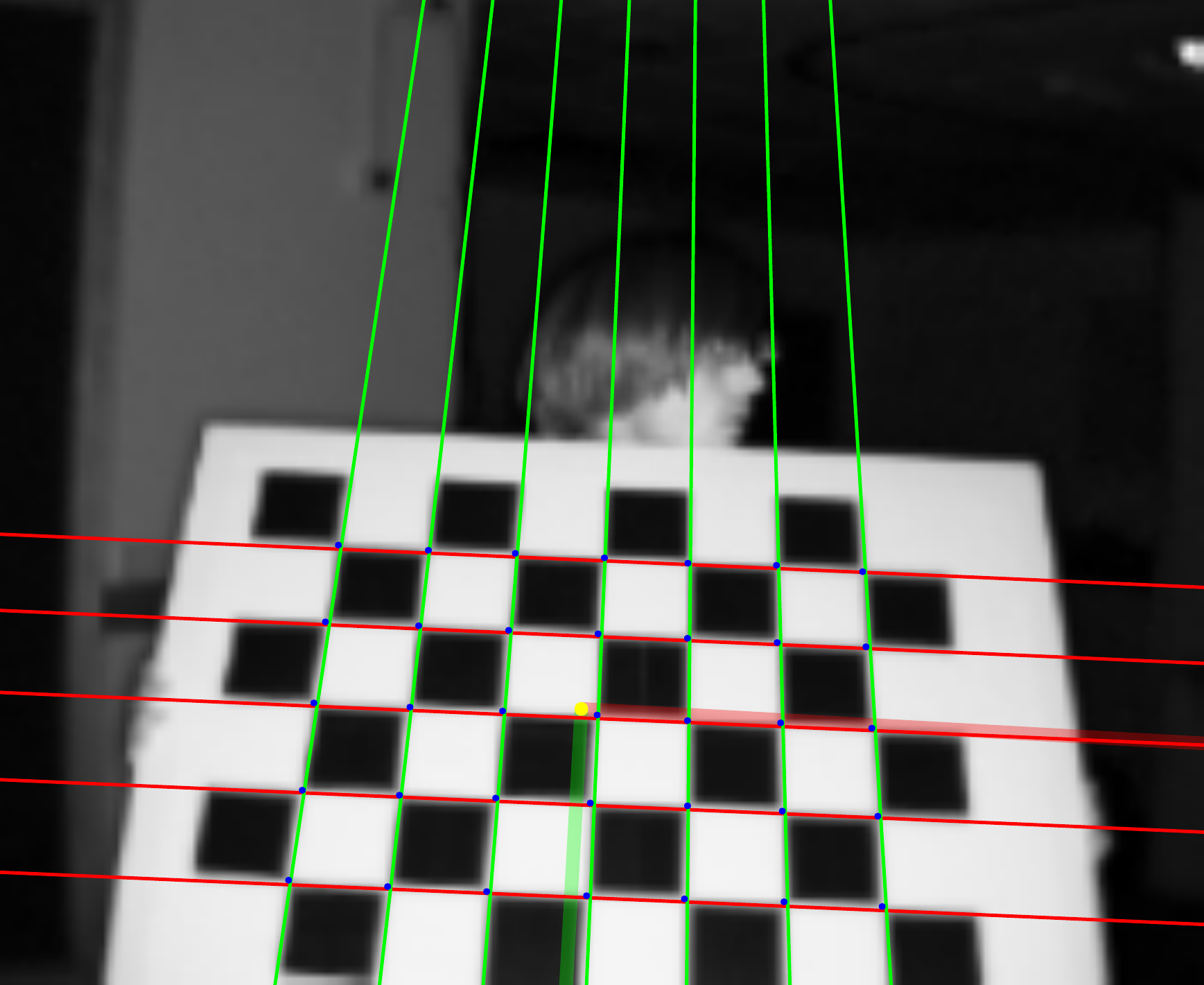}
\caption{Example detections (cf.\ fig.~\ref{fig:boards-1}) showing 
significant perspective effects.}
\label{fig:boards-2}
\end{figure}

\begin{figure}[ht!]
\centering
\includegraphics[width=.49\linewidth]{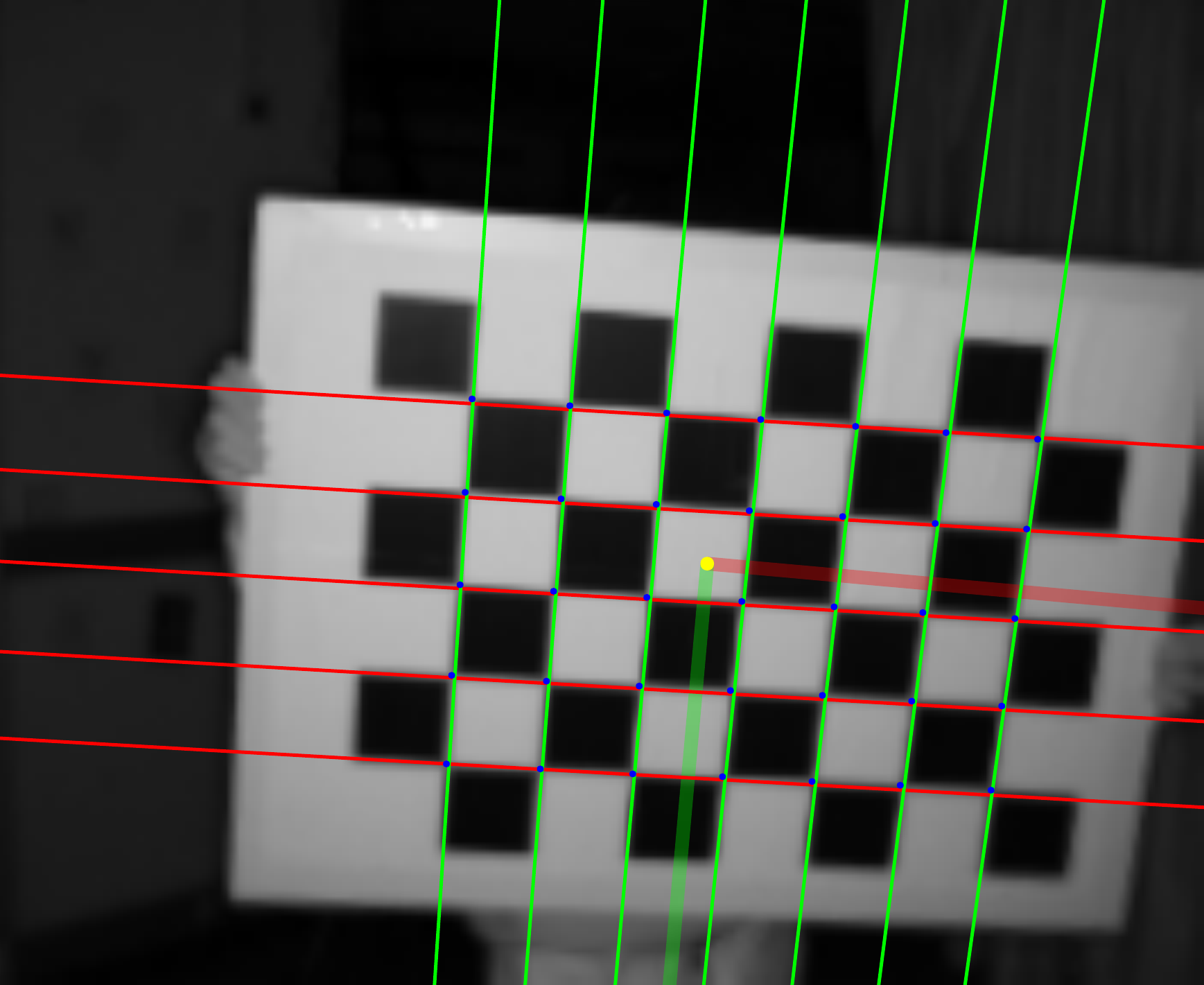}
\includegraphics[width=.49\linewidth]{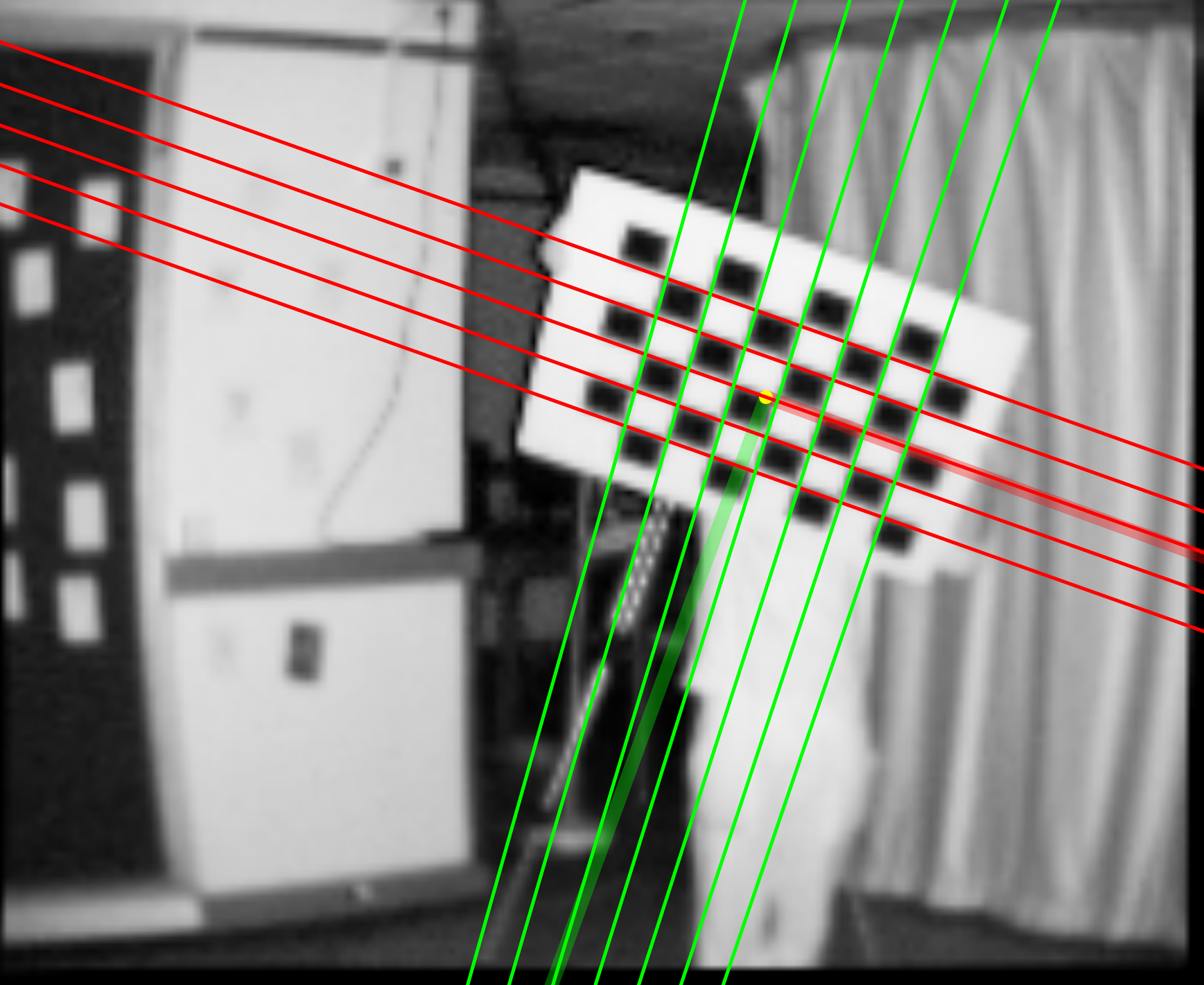}
\caption{Example detections (cf.\ fig.~\ref{fig:boards-1}) showing 
significant scale changes. The board on the right, which is in an 
image that shows background clutter and lens distortion, was not 
detected by OpenCV.}
\label{fig:boards-3}
\end{figure}

There were zero false-positive detections (100\% precision), 
as explained in sec.~\ref{sec:hough-analysis}. The number of
true-negatives is not useful here, because it depends largely 
on the configuration of the cameras (i.e.\ how many images show 
the back of the board). The false-negatives do not provide a 
very useful measure either, because the definition of these would depend on an arbitrary 
judgement about which of the very foreshortened boards `ought' 
to have been detected (i.e.\ whether an edge-on board is `in' the image or not).
It is emphasized that the test-data are actual calibration sets, containing many problematic images, and so the evaluation is based in a real-world application.
Some example detections are shown in figs.~\ref{fig:boards-1}--\ref{fig:boards-3},
including a variety of difficult cases.


\section{Discussion}
\label{sec:discussion}

A new method for the automatic detection of calibration grids in time-of-flight
images has been described.
The method is based on careful reasoning about the global geometric structure of the
board, before and after perspective projection. The method detects many more 
boards than existing heuristic approaches, which results in a larger and
more complete data-set for subsequent calibration algorithms. This is achieved while also \emph{reducing} the overall geometric error.
The increased number and variety of detections is of great benefit to subsequent extrinsic calibration procedures.


\subsection{Future work}

The Hough transform was developed, in section~\ref{sec:hough-transform}, as a dense \twod\ array. This presentation has the advantage of making the method easy to visualize and implement. However, it also raises issues of resolution and scalability \cite{zhang-1996,guo-1999}. In particular, the implementation in section~\ref{sec:hough-transform} is inefficient, both in space and time. These issues could be addressed by the use of a \emph{randomized} Hough transform \cite{kalviainen-1995}. 
This approach avoids building a dense transform array, in favour of a dynamic data structure~\cite{xu-1990}. Future work will examine the advantages of randomized methods, in relation to the increased complexity of implementation.
Another possible direction for future work would be to perform a global
refinement of the line-pencils, in the geometric parameterization, but
by minimizing a photometric cost-function with respect to the original images.

In a more general view, the present work suggests that it would be useful to have a data-set of calibration images with known 3D poses. These could, for example, be acquired using a robotic mounting of the physical board. This would enable a fine-grained comparison of algorithms, thereby encouraging future work in this area.

\bibliographystyle{elsarticle-num}
\bibliography{findboard}






\end{document}